\documentclass[reprint,onecolumn,amsmath,amssymb,jap,aip]{revtex4-1}
\pdfoutput=1 
\usepackage{graphicx}
\usepackage{dcolumn}
\usepackage{bm}
\usepackage{epsfig}
\usepackage{mathptmx}
\usepackage{times}
\usepackage{amsmath}
\usepackage{amssymb}
\usepackage{dcolumn}
\usepackage{units}
\usepackage{upgreek}
\usepackage{tabularx}
\usepackage{threeparttable}
\usepackage[space]{grffile}
\usepackage[normalem]{ulem}

\usepackage{subfig}
\usepackage{multirow}
\usepackage{array}
\usepackage{soul}
\usepackage{svg}
\usepackage{bbm}
\usepackage{verbatim}
\usepackage{multirow}
\usepackage{booktabs}
\usepackage{stmaryrd}
\usepackage{placeins}
\usepackage{hyperref}
\DeclareMathOperator*{\argmax}{argmax}
\newcommand{\Binary}{\lbrace 0,1\rbrace}
\newcommand{\Bipolar}{\lbrace -1,+1\rbrace}
\usepackage{xspace}
\newcommand{\name}{NVSA\xspace}
	
\newcommand{\cc}[1]{\normalsize{\color{black}#1}}
\newcommand{\ccc}[1]{\normalsize{\color{black}#1}}

\newcolumntype{C}{>{$}c<{$}}
\AtBeginDocument{
\heavyrulewidth=.08em
\lightrulewidth=.05em
\cmidrulewidth=.03em
\belowrulesep=.65ex
\belowbottomsep=0pt
\aboverulesep=.4ex
\abovetopsep=0pt
\cmidrulesep=\doublerulesep
\cmidrulekern=.5em
\defaultaddspace=.5em
}

\begin{document}
\title{A Neuro-vector-symbolic Architecture for Solving Raven’s Progressive Matrices}

\author{Michael Hersche} \affiliation{IBM Research -- Zurich, S\"{a}umerstrasse 4, 8803 R\"{u}schlikon, Switzerland.}\affiliation{Department of Information Technology and Electrical Engineering, ETH Z\"{u}rich, Gloriastrasse 35, 8092 Z\"{u}rich, Switzerland.}
\author{Mustafa Zeqiri} \affiliation{Department of Information Technology and Electrical Engineering, ETH Z\"{u}rich, Gloriastrasse 35, 8092 Z\"{u}rich, Switzerland.}
\author{Luca Benini} \affiliation{Department of Information Technology and Electrical Engineering, ETH Z\"{u}rich, Gloriastrasse 35, 8092 Z\"{u}rich, Switzerland.}
\author{Abu Sebastian} \email{ase@zurich.ibm.com} \affiliation{IBM Research -- Zurich, S\"{a}umerstrasse 4, 8803 R\"{u}schlikon, Switzerland.}
\author{Abbas Rahimi} \email{abr@zurich.ibm.com} \affiliation{IBM Research -- Zurich, S\"{a}umerstrasse 4, 8803 R\"{u}schlikon, Switzerland.}


\begin{abstract}
Neither deep neural networks nor symbolic AI alone has approached the kind of intelligence expressed in humans. This is mainly because neural networks are not able to decompose joint representations to obtain distinct objects (the so-called binding problem), while symbolic AI suffers from exhaustive rule searches, among other problems. These two problems are still pronounced in neuro-symbolic AI which aims to combine the best of the two paradigms. Here, we show that the two problems can be addressed with our proposed neuro-vector-symbolic architecture (\name) by exploiting its powerful operators on high-dimensional distributed representations that serve as a common language between neural networks and symbolic AI. The efficacy of \name is demonstrated by solving the Raven's progressive matrices datasets. Compared to state-of-the-art deep neural network and neuro-symbolic approaches, end-to-end training of \name achieves a new record of 87.7\% average accuracy in RAVEN, and 88.1\% in I-RAVEN datasets. Moreover, compared to the symbolic reasoning within the neuro-symbolic approaches, the probabilistic reasoning of \name with less expensive operations on the distributed representations is two orders of magnitude faster. Our
code is available at \url{https://github.com/IBM/neuro-vector-symbolic-architectures}.
\end{abstract}
\maketitle

\section*{}
%
Human fluid intelligence is the ability to think and reason abstractly, and make inferences in a novel domain.
The Raven’s progressive matrices (RPM)~\cite{Raven1938} test has been a widely-used assessment of fluid intelligence and abstract reasoning~\cite{Carpenter1990,RavenTest2012}.
The RPM is a non-verbal test which involves perceiving pattern continuation, element abstraction, and finding relations between abstract elements based on underlying rules.
Each RPM test is an analogy problem presented as a 3×3 pictorial matrix of context panels. 
Every panel in the matrix is filled with several geometric objects based on a certain rule, except the last panel which is left blank.
The participants are asked to complete the missing panel in the matrix by picking the correct answer from a set of candidate answer panels that matches the implicit rule (see Methods and Supplementary Fig.~1c).
Solving this test mainly involves two aspects of intelligence: visual perception and abstract reasoning.

How perception is combined with reasoning, and how they interact vary greatly across the spectrum of AI architectures. 
At one end of the spectrum, in the deep learning architectures~\cite{WReN,zheng2019abstract,CoPINET_19,Raven_19,I-Raven,MultiLayerRel_IJCNN2020,MRNet_CVPR2021,RPM_ImageNet_pretraining_2020,RPM_unsupervised} perception has primacy, and reasoning is more likely to adapt to the representations than vice versa.
At the opposite ends of the spectrum, in the classical symbolic AI, the perceptual representations are pre-engineered to e.g., emphasize relations, rather than the representations being generated as a consequence of the reasoning processes. 
In fact, it has been argued that the construction of appropriate representations is part of the reasoning process~\cite{Chalmers_1992}. 
Neuro-symbolic architectures further enrich the spectrum of possibilities by utilizing techniques from both ends: they combine subsymbolic (e.g., neural network) with symbolic AI approaches, aiming to reach human-level generalization~\cite{FODOR_1988,Neural-symbolic_Book_2002,AlgebraicMind_2001,Marcus_2020}. 
Considerable effort has been devoted to integrate the two ends that led to state-of-the-art performance of neuro-symbolic architectures in various tasks, e.g., visual question answering~\cite{NS-VQA_NIPS18,NS_ConceptLearner_ICLR19,NS_MetaConcept_NIPS19,Falcon_ICLR2022}, causal video reasoning~\cite{CLEVRER_ICLR2020}, and solving RPM~\cite{PrAE_CVPR21,nesy2022_knowledge}.
However, the resulting neuro-symbolic architectures are not immune to the potential problems of their individual ingredients (i.e., the neuro and symbolic parts), which are explained in the following.

%
We first explain the binding problem~\cite{Rosenblatt_1961} in neural networks that refers to their inability to recover distinct objects from their joint representation.
This inability prevents the neural networks from providing an adequate description of real-world objects or situations that can be represented by hierarchical and nested compositional structures~\cite{RachkovskijBinding2001}.
When we consider the fully local representations, an item of any complexity level can be represented by a single unit, e.g., by one-hot code. 
Such local representations limit the number of representable items to the number of available units in the pool, and hence cannot represent the combinatorial variety of real-world objects.
To address this issue, the distributed representations can provide enough capacity to represent a combinatorially growing number of compositional items.
However, they face another issue known as ``superposition catastrophe''~\cite{MalsburgAssemblies1986,SupCatas1999}.
Let us consider four atomic items \texttt{red}, \texttt{blue}, \texttt{square}, and \texttt{triangle}.
For representing two composite objects, e.g., a \texttt{red square} and a \texttt{blue triangle}, the activity patterns corresponding to their atomic items are superimposed without increasing the dimensionality.
As shown in Fig.~\ref{fig:binding}a, this results in a blend pattern that is ambiguous because the patterns of all four atomic items are coactivated, which causes ``ghost'' or ``spurious'' memory for unavailable objects such as \texttt{red triangle} or \texttt{blue square}.
As a common practice, to bypass this problem, the neuro-symbolic architectures employ two complementary networks (e.g., a Mask R-CNN followed by a ResNet-34) to be able to unambiguously extract the item attributes from multiple objects in an image~\cite{NS-VQA_NIPS18,NS_ConceptLearner_ICLR19,NS_MetaConcept_NIPS19,Falcon_ICLR2022}. 
The first network (Mask R-CNN) generates segment proposals for all objects such that each object can be processed individually by another network. 
Next, the generated bounding box for each single object paired with the original image is sent to the second network (ResNet-34) for extracting all attributes.
However, this approach of having multiple networks increases the number of weights.

%
The second ingredient of the neuro-symbolic architectures is the symbolic engine which is used for logical reasoning.
For solving the RPM tests, the symbolic logical reasoning is implemented as a probabilistic abduction reasoning~\cite{PrAE_CVPR21} that can be viewed as searching for a solution in a space defined by prior background knowledge.
The background knowledge about the rules is represented in symbolic form by describing all possible rule realizations that could govern the RPM tests~\cite{Carpenter1990}.
Then, the reasoning process can be solved via an exhaustive search over these symbols to abduce the probability distribution of the rules.
The computational complexity of the exhaustive search rapidly increases with the number of objects in the RPM panels. 
In effect, this exhaustive search problem hinders the utilization of the symbolic logical reasoning for end-to-end training and real-time inference.
%

%
We aim to address the two aforementioned problems by exploiting vector-symbolic architectures (VSAs)~\cite{MAP_1998,VSA_03,PlateHolographic1995,PlateHolographic2003,HD_09}.
VSAs are computational models that rely on high-dimensional distributed vectors and algebraic properties of their powerful operations to incorporate the advantages of connectionist distributed representations as well as structured symbolic representations (See~\cite{VSA_Survey_Part1,VSA_Survey_Part2} for review).
In a VSA, all representations---from atomic to composite structures---are high-dimensional holographic vectors of the same, fixed dimensionality.
The VSA representations can be composed, decomposed, probed, and transformed in various ways using a set of well-defined operations, including binding, unbinding, bundling (i.e., additive superposition), permutations, inverse permutations, and associative memory (i.e., cleanup).
Such characteristics of compositionality and transparency enable the use of VSAs in analogical reasoning~\cite{GreatSymbols_1998,AnalRetrieval_2000,AnalMapp_2009,RasmussenInductiveReasoning2011,EmruliAnalogical2013}, but these inspiring works do not have any perception module to process the raw sensory inputs. 
Instead, they assume there would be a perception system, e.g., a symbolic parser, providing the symbolic representations that support the reasoning.
Further, their inductive logical reasoning has eluded
the uncertainty in perception, and supported only one type of rules in RPM: the progression rule~\cite{RasmussenInductiveReasoning2011,EmruliAnalogical2013}.

Here, we propose a neuro-vector-symbolic architecture (\name) in which two powerful machineries, namely, deep neural networks and VSAs are synergistically combined to codesign a visual perception frontend and a probabilistic reasoning backend. Hereby, the demands of reasoning can drive the construction of appropriate perceptual representations. This synergy permits both perception and reasoning ends to tap into the rich resources of VSA as a general computing framework to overcome the problems of neural binding and exhaustive symbolic search as mentioned above, and yet they can be trained end-to-end. Accordingly, the main design objective of the \name frontend is to transduce the visual raw sensory inputs to the fixed-width VSA representations (as nested composable structures shown in Fig.~\ref{fig:binding}b). It is achieved by training nonlinear transformations of a neural network as a flexible means of representation learning over VSA. The resulting \name frontend addresses the binding problem in the neural networks, especially the superposition catastrophe, by effectively mapping the raw image of multiple objects to the structural VSA representations that still maintain the perceptual uncertainty. These representations can be readily used to solve visual analogy tasks by directly applying binding operations, or can be used to infer the probability mass functions per individual object attributes for further reasoning processes in the backend. The \name backend maps the inferred probability mass functions into another vector space of VSA such that the exhaustive probability computations and searches can be substituted by algebraic operations in that vector space. The design objective of the \name backend is to facilitate a differentiable and computationally efficient probabilistic reasoning process to support a set of RPM rules. The VSA operations offer distributivity and computing-in-superposition, which significantly reduce the computational costs thus performing probabilistic abduction and execution to predict the missing RPM panel in a generative and real-time manner.

\section{\name frontend: Perception}
Prior to describing the NVSA frontend, let us provide a brief background on VSA.
VSA is a family of similar systems that represent data using random high-dimensional vectors (see~\cite{VSA_Survey_Part1} for a review).
In this section, we use VSA whose vector entries are restricted to being dense bipolar~\cite{VSA_03}.
Initially, one or multiple codebooks are defined as $X :=\lbrace \mathbf{x}_i\rbrace_{i=1}^{m}$, where the elements of each atomic $d$-dimensional vector $\mathbf{x}_i \in \Bipolar^d$ are randomly drawn from a Rademacher distribution (i.e., equal chance of elements being ``-1'' or ``+1'').
We compare two vectors using the cosine similarity ($\mathrm{sim}$).
The similarity between two atomic vectors is close to zero with a high probability when $d$ is sufficiently large, typically in the order of thousands~\cite{HD_09}; hence, all vectors in the codebooks are quasi-orthogonal with respect to each other.

For representing a given data structure, VSA provides a set of well-defined vector operations. 
\emph{Bundling} ($\oplus$), superposition, or addition of two or more atomic vectors is defined as the element-wise sum with a subsequent bipolarization operation that sets the negative elements to \mbox{``-1''} and the positive ones to \mbox{``+1''}. 
This operation preserves similarity. 
On the other hand, \emph{binding} ($\odot$), or multiplication, of two or more vectors is defined with their element-wise product. 
Binding yields a vector that is dissimilar to its atomic input vectors. 
Every vector $\mathbf{x}_i$ $\in X$ is its own inverse with respect to the binding operation, i.e., $\mathbf{x}_i \odot \mathbf{x}_i =\mathbf{1}$, where $\mathbf{1}$ is the $d$-dimensional all 1-vector. 
Hence, the individual factors in a product can be exactly retrieved by \emph{unbinding}: $
\mathbf{x}_i \odot \left(\mathbf{x}_i \odot \mathbf{x}_j\right) =\left( \mathbf{x}_i \odot \mathbf{x}_i\right)  \odot \mathbf{x}_j = \mathbf{x}_j.$

The \name frontend consists of a trainable neural network and a VSA machinery whose algebra allows construction of appropriate perceptual representations suitable for the cognitive demands. 
The design of frontend is inspired by the expressiveness of high-dimensional VSA representations as a \emph{common language} between the symbolic representation of a set of objects and their data-driven representations obtained from a neural network. 
On the one hand, the real-world objects with arbitrary complexities can be symbolically described by VSA, e.g., a scene with multiple objects can be expressed as a hierarchical nested compositional structure of attributes, objects, and a set of objects---all with fixed-width vector arithmetic as shown in Fig.~\ref{fig:binding}b.   
On the other hand, the resulting VSA representations can also serve as an interface to a data-driven neural network whose layers of transformation are trained accordingly to transduce the raw image of the scene to the VSA representations (see Fig.~\ref{fig:NVSA}b). 
This novel combination merges the data-driven representation of the scene into its descriptive symbolic representation for facilitating down-stream reasoning tasks, without exploding the representation dimensionality, or facing the superposition catastrophe.
In the following, we describe the steps involved in the \name frontend.

\subsection{Defining VSA representations}
\begin{figure}[!ht]
\centering
\includegraphics[width=1.\textwidth]{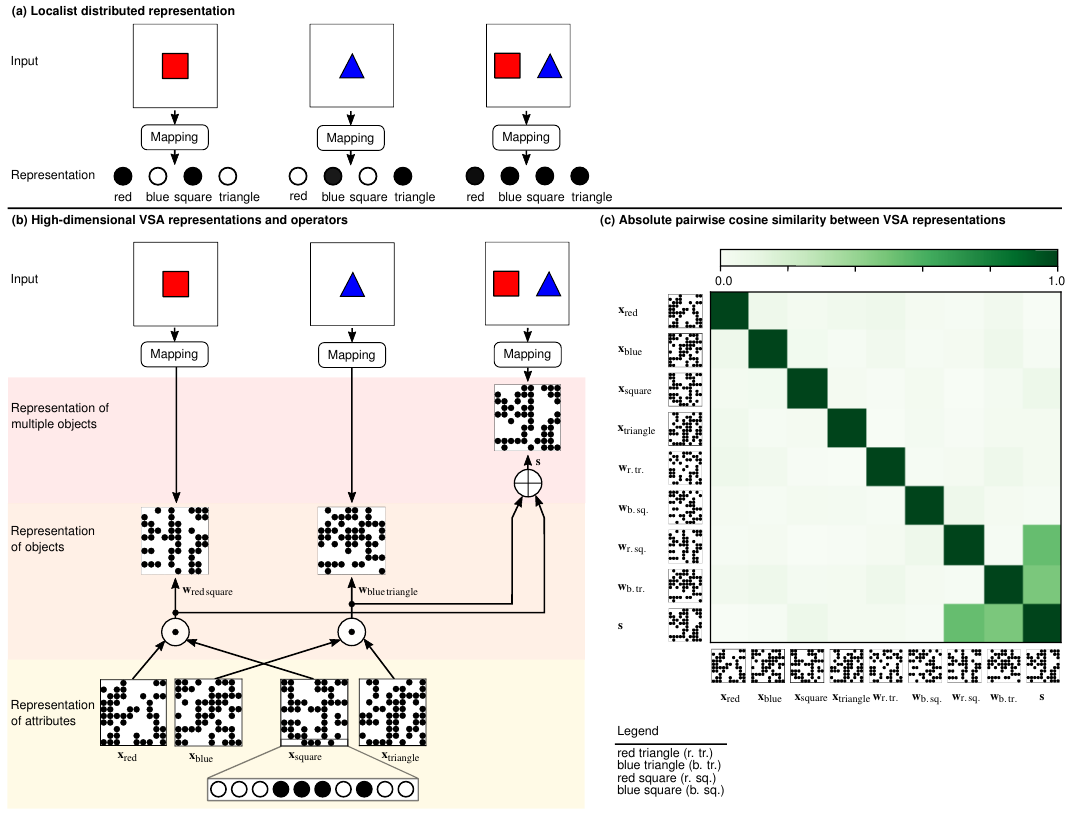}
\caption{\textbf{Illustration of the binding problem in the neural networks and our solution.}
\textbf{(a)} A localist distributed representation is used for objects with the color and shape attributes. The color attribute is mapped locally on a group of two neurons (\texttt{red} vs. \texttt{blue}). The shape attribute is similarly mapped to \texttt{square} vs. \texttt{triangle} neurons using a trainable mapping. This results in two distinct activated patterns for a \texttt{red square} and a \texttt{blue triangle}. If one of these objects are presented, the responses of the output neurons are sufficient to determine the identity of the object (i.e., its color and shape). However, when both objects are presented, their elementary patterns (symbols) are simultaneously activated that leads to binding ambiguity meaning that the resulting blend activity is insufficient to determine which object is in which color. This is often referred to as the superposition catastrophe~\cite{SupCatas1999}.
\textbf{(b)} The high-dimensional distributed VSA representations and operators can address this problem when properly combined with a neural network as the trainable mapping function. At the lowest level of the hierarchy, the four attribute values are represented by randomly drawing four $d$-dimensional vectors ($\mathbf{x}_\mathrm{red}, ...$). The vectors are dense binary, and arranged as $d=10\times10$ for the sake of visual illustration. At the next level, the \texttt{red square} object is described as a fixed-width product vector by binding two corresponding vectors ($\mathbf{x}_\mathrm{red} \odot \mathbf{x}_\mathrm{square}$) whose similarity is nearly zero to all attribute vectors and other possible product vectors such as $\mathbf{x}_\mathrm{blue} \odot \mathbf{x}_\mathrm{triangle}$, $\mathbf{x}_\mathrm{red} \odot \mathbf{x}_\mathrm{triangle}$, etc. as shown in \textbf{(c)}. This quasi-orthogonality allows the VSA representations to be co-activated with minimal interference. At the highest level, the two object vectors are bundled together by similarity-preserving bundling to describe the scene. The bundled vector is similar solely to those objects vectors and dissimilar to others.}
\label{fig:binding}
\end{figure}

This first step is to define a dictionary whereby the atomic concepts, their hierarchical compositions, and their relations can be described using the fixed-width vectors. 
What these concepts are can be guided by the cognitive demands. 
For instance, in solving RPMs, the reasoning process requires the probability mass functions of the object attributes, therefore the dictionary should be able to provide such appropriate representations by expressing attributes per objects which, in turn, fulfills the reasoning demand.    
The construction of the dictionary can be done by the application of the VSA operations on the VSA-encoded concepts as explained in the following. 
Let us consider the object attributes as the atomic concepts.
The encoding process starts by randomly generating a set of codebooks for the attributes of interest, e.g., one codebook for the colors ($\mathbf{x}_\mathrm{red}$, $\mathbf{x}_\mathrm{blue}$), and another codebook for the shapes ($\mathbf{x}_\mathrm{square}$, $\mathbf{x}_\mathrm{triangle}$).
Each codebook contains as many atomic $d$-dimensional vectors as there are attribute values.
It therefore provides a symbolic meaning for individual atomic vectors.
For describing an object with these two attributes, a product vector $\mathbf{w}$ can be computed by binding two vectors, one drawn from each of the codebooks (see Fig.~\ref{fig:binding}b).
For example, a \texttt{red square} object is represented as $\mathbf{w} = \mathbf{x}_\mathrm{red} \odot \mathbf{x}_\mathrm{square}$.
The random construction of the atomic vectors and the properties of multiplicative binding yield a unique $\mathbf{w}$ that is quasi-orthogonal (i.e., dissimilar) to the VSA representations of all other attributes and their combinations  (e.g., $\mathbf{x}_\mathrm{blue} \odot \mathbf{x}_\mathrm{triangle}$, or $\mathbf{x}_\mathrm{blue} \odot \mathbf{x}_\mathrm{square}$, or  $\mathbf{x}_\mathrm{red} \odot \mathbf{x}_\mathrm{triangle}$).
This means that the expected cosine similarity between two different object vectors is approximately zero with a high probability.
Therefore, when their VSA representations are coactivated, it results in minimal interference such that each object can be recovered (see Fig.~\ref{fig:binding}c.)

We have shown how to derive the object vectors from the elementary attribute vectors.
In the next level of the hierarchy, we are interested in an object-centric definition of the scene.
Here, we define the scene as the union of the objects.
Therefore, a scene with multiple objects is represented by bundling together their object vectors: $\mathbf{s}=(\mathbf{x}_\mathrm{red} \odot \mathbf{x}_\mathrm{square}) \oplus (\mathbf{x}_\mathrm{blue} \odot \mathbf{x}_\mathrm{triangle})$.
The bundling operation creates an equally-weighted superposition of multiple objects, and preserves similarity; hence, the bundled vector $\mathbf{s}$ is similar to both object vectors present in the scene, and dissimilar to other vectors in the system, as shown in Fig.~\ref{fig:binding}c.
This similarity preservation property allows the bundled vector to be solely matched with its constituent object vectors, which avoids the superposition catastrophe by design (e.g., $\mathrm{sim}(\mathbf{s}, \mathbf{x}_\mathrm{red} \odot \mathbf{x}_\mathrm{triangle})\approx0$). 
In summary, VSA can construct higher-level symbols of multiple objects by combining lower-level symbols of individual objects, and more elementary symbols of object attributes by using its dimensionality-preserving operators.

Next, we illustrate the generalization of the previous 2-attribute to 4-attribute objects suitable for solving RPM (an RPM test example is shown in Fig.~\ref{fig:NVSA}a). 
Similarly, we randomly generate a set of compact codebooks for the available attributes in the RAVEN dataset as $T :=\lbrace \mathbf{t}_i\rbrace_{i=1}^{5}$, $S :=\lbrace \mathbf{s}_i\rbrace_{i=1}^{6}$, $C :=\lbrace \mathbf{c}_i\rbrace_{i=1}^{10}$, and $L :=\lbrace \mathbf{l}_i\rbrace_{i=1}^{22}$ which respectively represent the type, size, color, and position of a single object, considering the cross-configuration equivalent positions with the same proportions (see Methods for more details).
We set $d=512$ that is sufficiently large to supply the atomic quasi-orthogonal vectors for every attribute value, while it is at least one order of magnitude smaller than the number of all possible combinations of attribute values ($m=6600$) for a single object.
Using these four codebooks, a quasi-orthogonal vector for every possible combination of a single object is computed as the Hadamard product of its attribute vectors (i.e., a 4-way multiplicative binding).
%
These $d$-dimensional vectors are stored in a dictionary $\mathbf{W}\in \Bipolar^{m\times d}$ which contains all $m$ possible single object combinations.
%
An arbitrary set of these single object vectors can be further composed by the bundling operation to describe e.g., an RPM panel as the union of its distinct objects.

The dictionary $\mathbf{W}$ is generated once, based on the initialization of codebook vectors, and is kept frozen during training.
As the second step of the frontend, in the next subsection, we show how this dictionary can be connected to the data-driven representations of its objects.

\subsection{Neural network representation learning over VSA}
\label{sec:neural_training}
To avoid the pitfalls of pure symbolic approaches and the need for a symbolic parser, we exploit the deep neural network representation learning over the defined VSA representations ($\mathbf{W}$) such that an image panel $\mathbf{X} \in \mathbb{R}^{r\times r}$ with resolution $r$ can be transformed and matched to the corresponding VSA representations using a mapping $f_\theta$ with learnable parameters $\theta$.
To do so, we propose using a ResNet-18, motivated by its good performance~\cite{Raven_19}, and interface its fully connected layer to the dictionary $\mathbf{W}$ as shown in Fig.~\ref{fig:NVSA}b.
With this interface, the last fully connected layer has an output dimension $d=512$ to be able to search on $\mathbf{W}$.
We insert a hyperbolic tangent ($tanh$) activation at the output of ResNet-18 to guide its real-valued output towards the bipolar representation of $\mathbf{W}$. 
By exploiting the VSA principles, the ResNet-18 can learn to superpose multiple, say $k$, objects in the $d$-dimensional vector from which all the attributes of the compound objects can be reliably recovered by $\mathbf{W}$ without facing the superposition catastrophe. 
Alternatively, other neuro-symbolic architectures~\cite{NS-VQA_NIPS18,NS_ConceptLearner_ICLR19,NS_MetaConcept_NIPS19,Falcon_ICLR2022} require at least two separate neural networks that collectively increase the number of trainable parameters by $\approx6\times$ compared to our \name frontend.

The \name frontend and backend can be trained end-to-end as we show in the next section.
However, in a fully supervised setting in which the labels of the visual attributes are given, the \name frontend can be trained independent of the backend.
Let $\mathbf{w}_1,\mathbf{w}_2,...,\mathbf{w}_m$ be the quasi-orthogonal representations of the object classes within $\mathbf{W}$, where $m$ is the number of single-object combinations.
For an image panel $\mathbf{X}$ containing $k$ objects, with $k$ target indices $\lbrace y_i \rbrace_{i=1}^k $, the trainable parameters $\theta$ of ResNet-18 can be optimized.
The optimization maximizes the similarity between the output query $\mathbf{q}=f_{\theta}(\mathbf{X})$ of ResNet-18 and the bundled vector $\mathbf{w}_{y_1}\oplus ... \oplus \mathbf{w}_{y_k}$ using a novel additive cross-entropy loss together with batched gradient descent. 
We provide the details of this loss, and show its superiority compared to other loss functions and perception methods (see Supplementary Note~1a).

We also analyze the generalization of the \name frontend to unseen combinations of attribute values in a novel object. 
Although we observe that the frontend with the multiplicative binding cannot generalize to unseen combinations of the attribute values, we enhance it by a multiplicative-additive encoding that can generalize up to 72\% (see Supplementary Note~1b).
The multiplicative binding-based encoding however generalizes well to unseen combinations of multiple objects (see Supplementary Note~1c).
Importantly, the employed multiplicative encoding results in learning the powerful perceptual representations in readiness for solving high-level reasoning tasks, such as visual analogy.
We have shown that the predicted perceptual representations at the output of ResNet-18 can be directly manipulated by the binding operations to solve visual analogy tasks ($A:B :: \alpha:\beta$). 
In the studied task, we consider a source domain that shares one relation, or multiple relations, between its two sets of objects ($A:B$), and a target domain that shares the same relation(s) between its object sets ($\alpha:\beta$).
Binding the neural network representations obtained from the source domain allows to capture the relation(s) solely from a single example, that can be applied to novel circumstances in the target domain by another application of binding operation. %
See Supplementary Note~2.

\subsection{Inferring probability mass functions from data-driven VSA representations}
We describe the last step of the frontend here.
Given an RPM panel, the ResNet-18 generates a VSA query vector that can be decomposed into the constituent object vectors, each derived from a unique combination of the attributes.
The decomposition performs a matrix-vector multiplication between the normalized dictionary matrix $\mathbf{W}$ and the normalized query vector, $\mathbf{q}$, to obtain the cosine similarity scores $\mathbf{z}$.
Since the structure of the dictionary matrix is known, we can infer the attributes namely, position, color, size, and type from the detected indices. 
Based on the similarity scores, we derive the probability mass functions (PMFs) for every object attribute, which include $\mathbf{v}_{type}$, $\mathbf{v}_{size}$, and $\mathbf{v}_{color}$ by marginalizing all non-negative similarity scores over all attribute combinations and a consecutive scaled softmax activation per object attribute.
In addition, we derive the probability of whether an object is present at a given position in $\mathbf{v}_{exist}$. 
After inferring these PMFs of the object attributes, we infer the PMFs of the panel attributes.
We combine all object PMFs to five PMFs which represent the position, number, type, size, and color distribution of the entire panel. 
These PMFs are denoted by $P:= (\mathbf{p}_{pos},\mathbf{p}_{num}, \mathbf{p}_{type},\mathbf{p}_{size}, \mathbf{p}_{color})$. 
See Methods for more details.

\begin{figure}[!ht]
\centering
    \fontsize{7}{10}
    \selectfont
\includegraphics[width=1.0\textwidth]{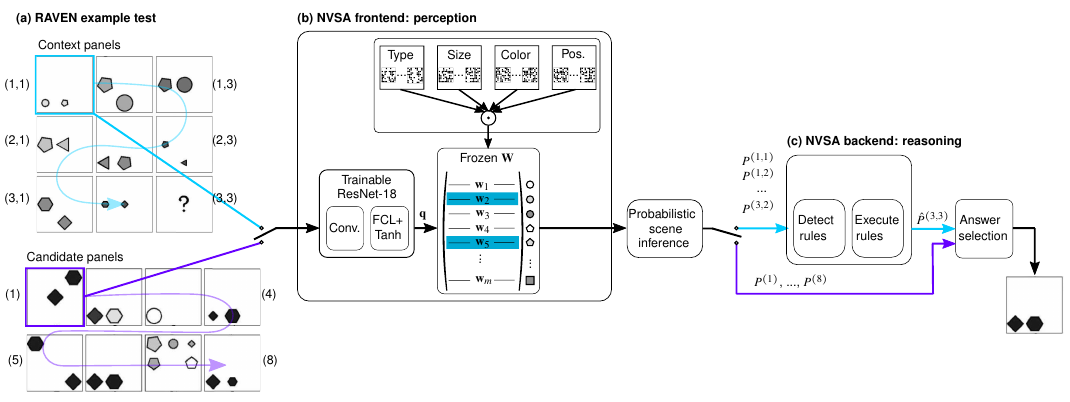}
\caption{\textbf{Proposed neuro-vector-symbolic architecture (\name).} \textbf{(a)} An example of RPM test taken from the RAVEN dataset that can be solved by \name. The context panels are composed of eight panels followed by a missing panel at (3,3). The candidate panels provide the potential candidates for the missing context panel. The correct answer is the 6th panel.
\textbf{(b)} \name frontend for perception. The frontend uses a trainable neural network (ResNet-18) and a frozen dictionary $\mathbf{W}$ that is generated by the four codebooks of $d$-dimensional vectors to cover all possible objects ($\mathbf{w}_1,\mathbf{w}_2,...,\mathbf{w}_m$, where $d\ll m$ in the RAVEN dataset).
The last fully connected layer of ResNet-18 with a $tanh$ activation is connected to $\mathbf{W}$. This dictionary forms a meaningful and semantically-informed VSA representation for individual object that is kept frozen during training, while the weights of ResNet-18 are learned by performing end-to-end training.
The training procedure directs ResNet-18 to generate a query vector, $\mathbf{q}$, such that it resembles the superposition of the VSA objects ($\mathbf{w}_2 \oplus \mathbf{w}_5$) available in the panel currently under the visual fixation. This results in merging the data-driven neural representation of the objects with their corresponding compositional VSA representation.  
\textbf{(c)} \name backend for reasoning. 
From the perceived similarities between $\mathbf{q}$ and $\mathbf{W}$, a probabilistic scene inference computes a set of probability mass functions (PMFs) for the attributes. The PMFs are shown as $P^{(i,j)}$ for every panel at location $(i,j)$ in the context matrix. These are used in the \name backend to predict the PMF of the missing panel ($\hat P^{(3,3)}$). The backend performs probabilistic calculations according to a set of rules in a differentiable and computationally-efficient manner, while not requiring any trainable parameters. The final answer is selected by choosing the candidate panel that minimizes the divergence between the predicted PMF ($\hat P^{(3,3)}$) and the PMF of the candidate panels ($P^{(1)}, ..., P^{(8)}$).
}
\label{fig:NVSA}
\end{figure}

Given an RPM test, we obtain a set of PMFs $P^{(i,j)}$ for each of the eight context panels, indexed by their row $i$ and column $j$, and a set of PMFs $P^{(i)}$ for each of the answer panels, as shown in Fig.~\ref{fig:NVSA}c. 
The set of context PMFs ($P^{(i,j)}$) form the probabilistic scene representations that are further transformed in the backend whose objective is to find the underlying rule.
The chosen rule is executed to generate $\hat P^{(3,3)}$ for the missing panel.

\section{\name backend: reasoning}
Here, we describe the \name backend that provides a computationally-efficient, differentiable, and transparent implementation of the probabilistic abductive reasoning.
The \name backend re-designs reasoning by exploiting the VSA representations and operators that permit handling large problem sizes that cannot be solved by traditional symbolic search-based reasoning approaches. 
As the first step in the \name backend, the inferred PMFs from the frontend are transformed into the distributed VSA representations in an appropriate vector space.
Next, this vector space should allow the application of VSA operators to implement the first-order logical rules such as addition of the attribute values, or subtraction, distribution, and more (see Fig.~\ref{fig:reasoning}a). 
The efficient VSA manipulations result in computing the rule probability for each possible rule, from which the most probable rule can be chosen and executed. 
These two main steps in the backend, followed by the end-to-end training of frontend and backend are described in the following subsections. 

\subsection{VSA representations of probability mass functions}
The RAVEN dataset~\cite{Raven_19} applies an individual rule to each of the five attributes (position, number, color, size, and type), which is either \texttt{constant}, \texttt{progression}, \texttt{arithmetic}, or \texttt{distribute three} (see Methods).
The rules are applied row-wise across the context matrix.
Based on the downstream rule, each attribute can be treated as continuous where there are relations among its set of values, or discrete where there are no explicit relations between the values.
For instance, the color attribute is treated as discrete in the \texttt{distribute three} rule, while the \texttt{arithmetic} rule treats it as continuous.
To make our VSA transformation general, we treat every attribute as both discrete and continuous, and it is up to the rule to use the proper representation.  
To achieve this in \name backend, we switch from the previously used bipolar dense representations to binary sparse block codes~\cite{laiho2015sparse,frady2021vfa}. 
This VSA framework with the help of fractional power encoding~\cite{PlateHolographic2003} permits the representation of continuous PMFs. 
The basis vectors in the binary sparse block codes are $d$-dimensional, binary-valued vectors with $\kappa$ non-zero elements. 
More specifically, the vectors are divided in $\kappa$ distinct blocks which contain exactly one non-zero element.  
The binding in the binary sparse block codes is defined as the block-wise circular convolution; similarly, the unbinding is the block-wise circular correlation. 
The similarity of two vectors is the sum the inner product normalized by the number of blocks $\kappa$. 
The bundling of two or more vectors is computed via the element-wise addition.
Optionally, the bundled vector could be sparsified to have only one non-zero element per block again, however, this results in loss of information. 
Hence, in this work, the bundling is performed without sparsification. 

In the following, we illustrate how a PMF can be transformed to this VSA format.
To represent the PMF of an attribute in the VSA space, we first generate a codebook $B:= \lbrace \mathbf{b}_i \rbrace_{i=1}^{n}$, where $\mathbf{b}_i \in \Binary ^ {d}$.
For a discrete attribute, we use a codebook with $n$ unrelated basis vectors $\mathbf{b}_i$. 
For representing the PMF of a continuous attribute, we use a codebook with basis vectors generated by the fractional power encoding~\cite{PlateHolographic2003}, where the basis vector corresponding to an attribute value $v$ is defined by exponentiation of a randomly chosen basis vector $\mathbf{e}$ using the value as the exponent, i.e., $\mathbf{b}_v = \mathbf{e}^v$.
See Methods on how to create the codebooks for discrete and continuous attributes.   
Each PMF is represented through the weighted superposition with the values in the PMF used as weights and the corresponding codewords as basis vectors (see Fig.~\ref{fig:reasoning}a):
\begin{align}\label{eq:block_superpos}
    \mathbf{a}^{(i,j)}:=g(\mathbf{p}^{(i,j)}) = \sum_{k=1}^{n} \mathbf{p}^{(i,j)}[k] \cdot \mathbf{b}_k,
\end{align}
Every attribute PMF is transformed separately to its corresponding VSA representation, e.g., the PMF of the attribute number is transformed to $\mathbf{a}_{num}^{(i,j)}:=g(\mathbf{p}_{num}^{(i,j)})$.

\subsection{VSA-based probabilistic abduction and execution}
\begin{figure}[!ht]
\centering
    \fontsize{7}{10}
    \selectfont
\includegraphics[width=0.9\textwidth]{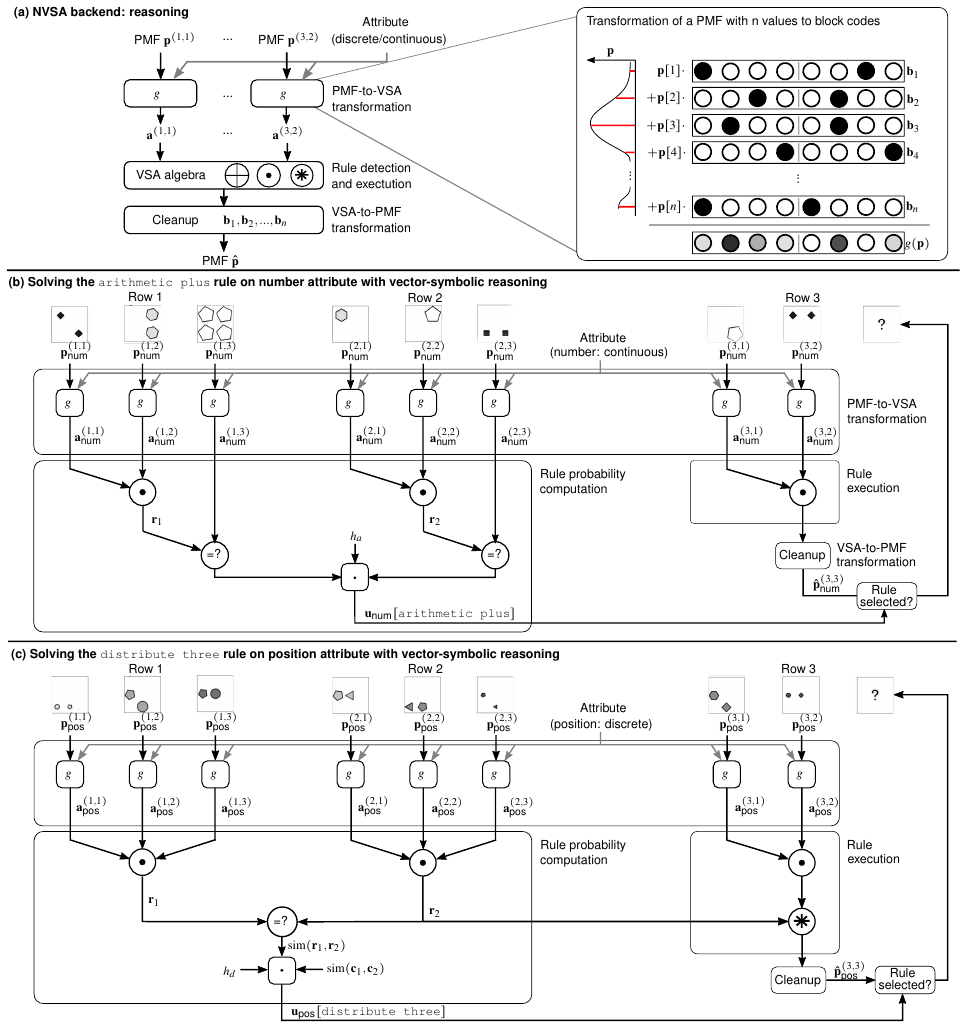}
\caption{\textbf{\name backend.} \textbf{(a)} Steps involved in the \name reasoning. First, a PMF of a discrete or a continuous attribute is transformed to the VSA format of binary sparse block codes. This transformation ($g$) is illustrated for the PMF of an attribute with $n$ possible values. Next, the VSA algebra can be applied on the VSA representations of PMFs to implement the rule of interest. Last, after executing the rule, the resulted VSA representation can be cleaned up by doing an associative memory search on the codebook of values that returns an output PMF. 
\textbf{(b)} The use of backend in solving the \texttt{arithmetic plus} rule. The PMF of each panel $\mathbf{p}_{num}^{(i,j)}$ is transformed to the VSA representation. In the rule probability computation step, the VSA representations of first two panels are bound together per row, yielding two row vectors $\mathbf{r}_1$ and $\mathbf{r}_2$. The rule probability is computed by multiplying the similarities between the row vectors $\mathbf{r}_i$ and the last panel of the first two rows, and an additional constraint $h_a$. In the rule execution, the VSA representation of the missing panel is predicted by binding the VSA representations of position (3,1) and (3,2). Finally, an associative memory cleanup computes the similarities between the predicted vector and all atomic vectors to determine the PMF $\mathbf{\hat p}_{num}^{(3,3)}$.
\textbf{(c)} The \name backend for solving the \texttt{distribute three} rule. In the rule detection step, the VSA representations of the first two rows are bound together per row, yielding $\mathbf{r}_1$ and $\mathbf{r}_2$. The rule probability is the product of the similarity between the two row vectors, the similarity between the first two column representations $\mathbf{c}_1$ and $\mathbf{c}_2$. In the rule execution, the VSA representation of the missing panel is predicted by unbinding the vector representations of position (3,1) and (3,2) from one of the two row representations, e.g., $\mathbf{r}_2$.
}  \label{fig:reasoning}
\end{figure}

The attribute PMFs of the panel are mapped to the VSA format where we can use the VSA algebra to implement the functions embedded in the underlying rules. 
Let us consider the \texttt{arithmetic plus} rule for the number attribute, which is treated as continuous and shown in Fig.~\ref{fig:reasoning}b.
In each row, the number of objects in the third panel is the sum of the number of objects in the first two panels.
As this rule is of continuous nature, we represent the PMFs using fractional power encoding.
The VSA representations of PMFs obtained in equation~\eqref{eq:block_superpos} are bound to compute $\mathbf{r}_{i}$ vectors for the first and second row using the first two panels: 
\begin{align}\label{eq:plus}
     \mathbf{r}^+_{i} = \mathbf{a}^{(i,1)} \odot \mathbf{a}^{(i,2)}, \quad i\in \lbrace 1,2 \rbrace. 
\end{align}

To better understand equation~\eqref{eq:plus}, let us assume that the distribution of the PMFs of the context panels is maximally compact, i.e., the values in $\mathbf{p}_{num}^{(i,j)}$ are ``1'' at the correct number of objects and ``0'' elsewhere.
Then, the bound vector of the first row can be formulated as $\mathbf{r}^+_{1}=\mathbf{e}^{v_1} \odot \mathbf{e}^{v_2} =  \mathbf{e}^{v_1+v_2}$, where $v_1$ and $v_2$ are the numbers of objects in the first and second panel. 
If the rule applies, i.e. $v_3 = v_1+v_2$, we expect the bound vector $\mathbf{r}^+_{1}$ to be identical to the VSA representation of the last panel in the row $\mathbf{a}^{(1,3)}=\mathbf{e}^{v_3}=\mathbf{e}^{v_1+v_2}$, thanks to the properties of fractional power encoding. 

For supporting arbitrary PMFs, we validate the rule using the similarity between the bound vectors.  
Combining the row-wise similarities with additional constraints yields us an estimation of the rule probability:
\begin{align} \label{eq:arithmetic}
    \mathbf{u}[\texttt{arithmetic plus}] = \mathrm{sim}( \mathbf{r}^+_{1},\mathbf{a}^{(1,3)} ) \cdot \mathrm{sim}( \mathbf{r}^+_{2},\mathbf{a}^{(2,3)} ) \cdot h_a(\mathbf{a}^{(3,1)},\mathbf{a}^{(3,2)}),
\end{align}
where $h_a$ is an additional rule-dependent constraint (see Supplementary Note~3). 
When the rule probability for the \texttt{arithmetic plus} is the highest among all possible rules, we estimate the vectorized representation of the number attribute for the missing panel by 
\begin{align}
    \mathbf{\hat{a}}^{(3,3)} = \mathbf{a}^{(3,1)} \odot \mathbf{a}^{(3,2)}. 
\end{align}
This bound vector represents the estimation of the PMF. 
If the PMFs in the last row are maximally compact, the bound vector corresponds to the correct number of objects of the missing panel.
Otherwise, the bound vector represents a superposition of the correct number vector and additional terms which can be considered as noise terms, stemming from the smaller non-zero contributions in the PMF. 

To compute the PMF of the missing panel attribute, we do an associative memory search between the bound vector and all atomic vectors in the codebook $B$, followed by a normalization: 
\begin{align} \label{eq:am}
    \mathbf{\hat p}_{num}^{(3,3)} =\mathrm{norm}\left(\left[ \mathrm{sim}\left(\mathbf{\hat a}^{(3,3)}, \mathbf{b}_1\right), \mathrm{sim}\left(\mathbf{\hat a}^{(3,3)}, \mathbf{b}_2\right), ..., \mathrm{sim}\left(\mathbf{\hat a}^{(3,3)}, \mathbf{b}_n\right)\right]\right).
\end{align}

Next, we show how the \name backend supports the rules with the discrete treatment of the attributes such as the \texttt{distribute three} rule. 
Without loss of generality, we explain our method for the position attribute in the panel constellation with a 3x3 grid (see Fig.~\ref{fig:reasoning}c).
The position is described with a 9-bit code where a ``1'' indicates that the position is occupied inside the 3x3 grid.
This 9-bit position index $p$ takes values from $1$ to $n=511$, considering the constraint of having at least one object per panel.
A different value of the position attribute (from 1 to $n$) appears in each of the three panels of a row. 
The \texttt{distribute three} requires that the same values appear in each row with a distinct permutation. 
The same holds with respect to the columns.

We transform the position PMF of every panel using equation~\eqref{eq:block_superpos} in combination with a discrete codebook $B$. 
These VSA representations are used to compute the product vectors for the first and second rows and columns of the context matrix:
\begin{align}
     \mathbf{r}_{i} &= \mathbf{a}^{(i,1)} \odot \mathbf{a}^{(i,2)} \odot \mathbf{a}^{(i,3)}, \label{eq:distthreedet_row} \\
     \mathbf{c}_{j} &= \mathbf{a}^{(1,j)} \odot \mathbf{a}^{(2,j)} \odot \mathbf{a}^{(3,j)}, \quad i,j\in \lbrace 1,2 \rbrace. \label{eq:distthreedet_col}
\end{align}
Equations~\eqref{eq:distthreedet_row} and~\eqref{eq:distthreedet_col} describe a VSA-based conjunctive formula grounded over the row and column being considered, respectively.
For example, given a set of arbitrary PMFs in a row, the resulting product vector ($\mathbf{r}_1$ or $\mathbf{r}_2$) is unique.
However, for any order of PMFs in the row, the computed product vectors are the same due to the commutative property of the binding operation. 
We exploit this property to detect whether the \texttt{distribute three} rule applies by simply checking if the product vectors are similar among rows and columns, i.e., $\mathrm{sim}(\mathbf{r}_1$,$\mathbf{r}_2)\gg 0$ and $\mathrm{sim}(\mathbf{c}_1$,$\mathbf{c}_2)\gg 0$, and combine them together to estimate the rule probability
\begin{align} \label{eq:dist3}
    \mathbf{u}[\texttt{distribute three}] = \mathrm{sim}(\mathbf{r}_1,\mathbf{r}_2)\cdot \mathrm{sim}(\mathbf{c}_1,\mathbf{c}_2) \cdot \cc{ h_d(\mathbf{a}^{(1,1)},\mathbf{a}^{(2,1)}, ...,\mathbf{a}^{(2,3)})}, 
\end{align}
where $h_d$ is an additional rule-dependent constraint (see Supplementary Note~3). 
To execute the rule, we first unbind two vectors ($\mathbf{a}^{(3,1)}$ and $\mathbf{a}^{(3,2)}$) from one of the row product vectors ($\mathbf{r}_1$ or $\mathbf{r}_2$), which results in an unbound vector $\mathbf{\hat a}^{(3,3)}$.
The PMF $\mathbf{\hat{p}}_{pos}^{(3,3)}$ of the missing panel is estimated by the associate memory in equation~\eqref{eq:am} which searches on the values of the position attribute.

The associative memory search is only limited to the $n$ atomic vectors in the codebook $B$; hence, our \name backend requires $\mathcal{O}(n)$ in time and space.
This is a significant reduction compared to pure symbolic search-based reasoning approaches which search exhaustively through all possible rule implementations that demanding up to $\mathcal{O}(n^3)$ in time and space. 
For example, the previously described \texttt{distribute three} rule on the attribute position would have $\binom{n=511}{3}\cdot12\geq 26\cdot10^7$ different rule implementations in the 3x3 grid constellation which is prohibitive to compute.
This exhaustive rule search forces the neuro-symbolic approach in~\cite{PrAE_CVPR21} to considerably limit its search space at the cost of lower accuracy. 
Instead, our approach efficiently covers the entire search space by simple binding and unbinding operations on the VSA representations followed by a linear associative memory search whose time and space complexity is set as the cube root of the exhaustive search space.
{\cc{
This computational advantage of VSA is mainly due to performing search-in-superposition. 
For example, by comparing the VSA representations of the first and second row, we can sum over all possible combinations in superposition: 
\begin{align}
    \mathrm{sim}(\mathbf{r}_1, \mathbf{r}_2) &=  \mathrm{sim}(\mathbf{a}^{(1,1)}\odot\mathbf{a}^{(1,2)} \odot \mathbf{a}^{(1,3)}, \mathbf{a}^{(2,1)}\odot\mathbf{a}^{(2,2)} \odot \mathbf{a}^{(2,3)}) \\
    &=\mathrm{sim}\left( \left(\sum_{k=1}^{n} \mathbf{p}^{(1,1)}[k]  \cdot \mathbf{b}_k\right) \odot \left(\sum_{k=1}^{n} \mathbf{p}^{(1,2)}[k]  \cdot \mathbf{b}_k\right) \odot  \left(\sum_{k=1}^{n} \mathbf{p}^{(1,3)}[k]  \cdot \mathbf{b}_k\right),  \mathbf{a}^{(2,1)}\odot\mathbf{a}^{(2,2)} \odot \mathbf{a}^{(2,3)}\right). 
\end{align}
Without the VSA representations, one would need to compute the maximally expanded version. 
}}
See Supplementary Note~3 for details about our implementation of the \texttt{artihmetic minus}, the \texttt{progression}, and the \texttt{constant} rule.

\subsection{End-to-end training}
We train our NVSA frontend and backend end-to-end.
Note that only the neural part of the frontend (i.e., ResNet-18) has trainable parameters, while the dictionary ($\mathbf{W}$) and all the parameters in the backend (e.g., rule representations) are frozen.
Every training RPM example provides eight context panels $(\mathbf{X}^{(1,1)},...,\mathbf{X}^{(3,2)})$, eight candidate answer panels $(\mathbf{X}^{(1)},...,\mathbf{X}^{(8)})$, the ground-truth answer $y$, and the ground-truth rule per attribute $\mathbf{r}$. 
First, we pass the context panels through the NVSA frontend and infer $P^{(i,j)}$; similarly, we infer $P^{(i)}$ for the candidate answer panels. 
Using $P^{(i,j)}$, we compute the rule belief per attribute using the NVSA backend (e.g., using equations~\eqref{eq:arithmetic} and \eqref{eq:dist3} for \texttt{arithmetic plus} and \texttt{distribute three}). 
Based on the distribution of the rule beliefs, we then sample an action per attribute and execute it, yielding five probability distributions for the attributes  $\hat{P}^{(3,3)}$.
Finally, we compute for each candidate answer panel $j$, the Jensen–Shannon divergence (JSD) between each of the five probability distributions in $P^{(k)}$ and $\hat{P}^{(3,3)}$, and sum the five JSD values to obtain a score for the answer panel $j$.

Inspired by PrAE~\cite{PrAE_CVPR21}, we mutually minimize a loss based on REINFORCE~\cite{williams1992reinforce} and an additional auxiliary loss.
The REINFORCE-based loss combines the negative cross-entropy on the scores, interpreted as a reward function, with the log-likelihood of the sampled action.
The auxiliary loss sums up the negative log-likelihood of the ground-truth rules (See Methods). 
Hence, by minimizing the auxiliary loss, we train the frontend to map the context panels to the PMFs based on which the rule detection yields the correct rule.

%

\section{Results}
\begin{figure}[!ht]
\centering
    \subfloat[RAVEN]{\includegraphics[width=.45\textwidth]{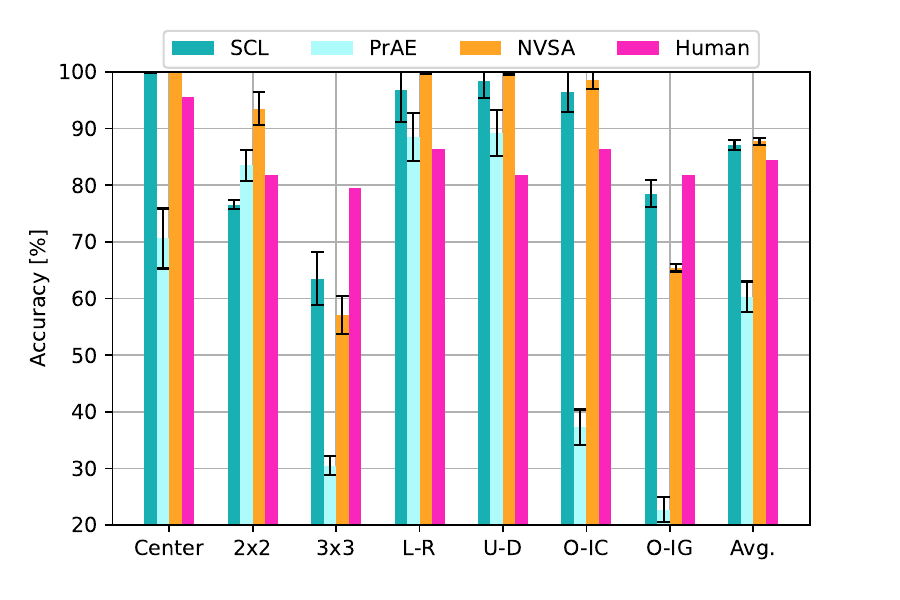}\label{fig:RAVEN}}
    \subfloat[I-RAVEN]{\includegraphics[width=.455\textwidth]{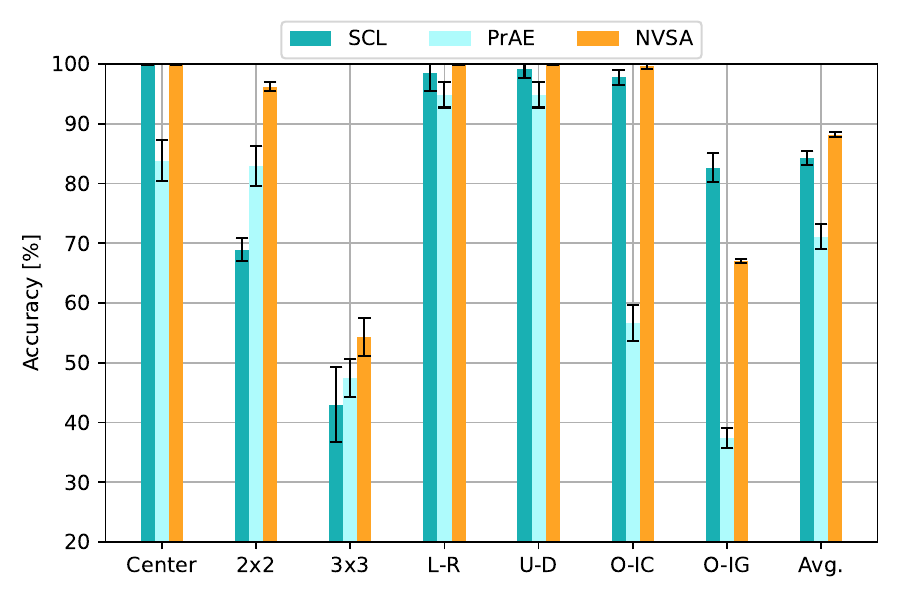}\label{fig:I-RAVEN}}
\caption{\textbf{Average accuracy on the (a) RAVEN and (b) I-RAVEN test sets.} It compares the classification accuracy of the state-of-the-art approaches in deep neural networks (SCL~\cite{wu2020scl}) and neuro-symbolic AI (PrAE~\cite{PrAE_CVPR21}) with our \name, when all models are trained end-to-end. Error bars indicate the standard deviation from five training and validation runs with different seeds (the exact numeric values are provided in the Extended Tables~{\ref{tab:results-overall-RAVEN} and \ref{tab:results-overall-IRAVEN}}). The human performance reported in~\cite{Raven_19} is also shown in \textbf{(a)} for the RAVEN dataset. The I-RAVEN dataset is a revised unbiased version of the RAVEN dataset which generates a fair set of answer panels to eliminate the shortcut solution in the RAVEN dataset (see Methods for details).} \label{fig:results}
\end{figure}

%
We evaluate \name on the RAVEN~\cite{Raven_19}, I-RAVEN~\cite{I-Raven}, and PGM~\cite{WReN} datasets (see Methods). 
First, we consider more diverse RAVEN and I-RAVEN datasets.
Fig.~\ref{fig:results} compares the classification accuracy with the state-of-the-art models in pure deep neural networks {\ccc{(SCL~\cite{wu2020scl})}} and neuro-symbolic AI (PrAE~\cite{PrAE_CVPR21}), \cc{where we have retrained both models five times using different random seeds and used the checkpoint with the highest accuracy on the validation set}.
A separate SCL model was trained per constellation.
On the RAVEN dataset, \name achieves an average accuracy of 87.7\%, outperforming SCL by 0.5\% and PrAE by 27.4\%. 

There is a short-cut solution in the answer panels of the RAVEN dataset (see Methods).
It has been shown that the shortcut pattern can be leveraged by deep neural networks, e.g., CoPINet~\cite{CoPINET_19} achieved a higher accuracy when being trained and tested exclusively on the answer panels without considering the context panels at all (context-blind)~\cite{I-Raven}.
In this regard, the understanding of the context matrix and its underlying rules is bypassed by shortcut learning from the answer set only. 
Therefore, it is recommended to use the I-RAVEN dataset~\cite{I-Raven}, which provides unbiased fair answer panels, when testing RPM reasoning models~\cite{RPM_Survey2022,Mitchel_Survey_2021}.
\name achieves the highest accuracy on the I-RAVEN dataset too (88.1\%) on average, while the majority of deep learning approaches~\cite{WReN,Raven_19,zheng2019abstract,CoPINET_19,zhuo2021dcnet} face a large accuracy drop by showing $<50\%$ accuracy on average.
Our \name does not face any accuracy drop by switching from RAVEN to I-RAVEN because it cannot rely on such a shortcut by design: based on the highest probable rules, it first makes a prediction of the PMFs of the empty panel before individually comparing it to each PMF of the answer panel.
The controllability and explainability of \name is a great advantage for problems that require it.
\name also significantly outperforms \ccc{SCL by 4.2\% and PrAE by 17.0\%} on I-RAVEN, on average.
Extended Tables~{\ref{tab:results-overall-RAVEN} and \ref{tab:results-overall-IRAVEN}} present a detailed comparison with the state-of-the-art methods in the tabular format.

Next, we compare the accuracy and the compute time of the \name backend with the PrAE reasoning backend by providing the ground-truth attribute values. 
As shown in Table~\ref{tab:results-reasoning}, the PrAE reasoning backend reaches relatively lower accuracies (94.21\%--95.68\%) in the 2x2 grid, the 3x3 grid, and the out-in grid compared to the other constellations.
We identify the root cause of the low accuracy in these three constellations to be the approximations made in the exhaustive search by applying restrictions to get faster execution. 
We remove these search restrictions from PrAE and create an unrestricted PrAE. 
This increases the accuracy of those three constellations to 97.5\%--99.22\%.
While the compute time of the unrestricted PrAE remains similar for most configurations, it increases rapidly for the 3x3 grid, requiring 15,408\,minutes (10.7\,days) instead of the previous 648\,minutes (10.8\,hours) in the PrAE with restricted search for solving 2000 RPM tests.
Note that we run the experiments on the CPUs as the unrestricted PrAE demands more than 53\,GB memory that could not fit the GPU providing 32\,GB memory (see Methods). 
However, our \name reasoning backend effectively resolves this bottleneck: it reduces the computation time on the 3x3 grid to 63.2\,minutes, which is 244$\times$ faster than the unrestricted PrAE, and the memory demand to $<10$\,GB, while maintaining the high accuracy (96.89\% vs. 97.50\%).
Moreover, we demonstrate that frontend and backend of our \name can be trained end-to-end, practically in any constellations, and it provides real-time inference for solving the RPM tests (see Supplementary Video).

Moreover, we also evaluate \name on the PGM dataset~\cite{WReN}, being the first neuro-symbolic approach targeting this dataset, while other neuro-symbolic works~\cite{PrAE_CVPR21,nesy2022_knowledge} only targeted RAVEN/I-RAVEN dataset. 
\name achieves an average accuracy of 68.3\% and is highly competitive with the reproduced state-of-the-art MRNet.
See Supplementary Note~5 for more details.

Lastly, we showcase the out-of-distribution generalizability of our \name with respect to unseen attribute-rule pairs in the I-RAVEN dataset. 
More specifically, we evaluate whether our model is able to solve tasks containing an unseen target attribute-rule pair (e.g., the constant rule on the type attribute) when it has been trained on the examples containing all of the attribute-rule pairs except the specific target one (e.g., the constant rule on size and color, the progression rule on all attributes, and the distribute rule on all attributes). 
Our \name outperforms the baselines (LEN~\cite{zheng2019abstract} and CoPINet~\cite{CoPINET_19}) by a large margin in all unseen attribute-rule pairs (See Supplementary Note~4). 

\section{Discussion}

The \name frontend allows expression of many more object combinations than the dimensions in the vector space. 
However, it requires to store and search on the dictionary $\mathbf{W}$.
Given the quasi-orthogonality of the representations in $\mathbf{W}$, it can be substituted with a set of smaller codebooks by potentially exploiting the VSA operators in a nonlinear dynamical system.
A powerful example of this would be the resonator networks~\cite{Resonator1,Resonator2} and their stochastic nonlinear variants~\cite{IMC_Factorizer} that can quickly factorize a product vector in an iterative manner thus reducing the computation/storage demand on the dictionary when decomposing an object vector.

The associative memory search is the central ingredient of \name in both perception and reasoning for estimating the PMFs.
To reduce the computational complexity of the associative memory, one notable option is to use in-memory computing that executes searches in an analog manner. 
It has recently been shown that the associative memory can be realized by analog in-memory computing based on crossbar arrays of emerging non-volatile memories~\cite{HDC_NatElec20,MANN_NatCom21,RRAM_1kb_HammDis,MemristiveSurvey}.
Besides improving the computational density and energy efficiency, this paves the way for reducing the timing complexity of the associative memory to $\mathcal{O}(1)$.
Other frequent primitives such as binding and bundling can also benefit from low-power hardware realization~\cite{semiHD_HW}. 

%
By accurately and efficiently solving RPM tests, we have demonstrated that \name enhances the aspects of both perception and reasoning by adding a distinctive vectorized flavor to them which is based on the high-dimensional distributed representations and operators of VSA.
In the proposed \name frontend, instead of naive local or distributed representations for the objects, we exploited high-dimensional VSA representations. 
A multi-attribute meaning was structurally assigned to every object vector by binding its attribute vectors, which can be further bundled to create a composite vector representing multiple objects---all in a fixed dimension that is significantly lower than the combinatorial attributes.
These structured representations were used as the target vectors to train the deep neural network.
The training can be done end-to-end, or by using the additive cross-entropy loss when the attribute labels are available. 
Being able to train this deep transformation permitted the simultaneous inference of multiple attributes of multiple objects in a visual scene with neither exploding the representation dimensionality, nor facing the superposition catastrophe.   
In the \name backend, we proposed a computationally-efficient and differentiable reasoning where the probability mass functions of discrete or continuous attributes are expressed as the VSA representations.
This permitted the use of VSA operators to efficiently implement the rules which save the computational cost significantly thanks to the distributivity and computing-in-superposition of VSA.
As a result, the time/space computational complexity of the \texttt{distribute three} rule search was reduced from $\mathcal{O}(n^3)$ to $\mathcal{O}(n)$, leading to two orders of magnitude shorter execution time.
It was shown that \name surpasses both pure deep learning~\cite{MRNet_CVPR2021} and neuro-symbolic~\cite{PrAE_CVPR21} state-of-the-arts by achieving average accuracy of 87.7\% in the RAVEN~\cite{Raven_19} and 88.1\% in the I-RAVEN~\cite{I-Raven} datasets.
\name also enabled real-time execution on CPUs, which is 244$\times$ faster than the functionally-equivalent symbolic logical reasoning.

\name is a significant step towards encapsulating different AI paradigms in a unified framework to address task involving both perception and higher-level reasoning.


\clearpage
\section*{Methods}

\subsection*{RAVEN and I-RAVEN dataset containing RPM tests}
The RAVEN dataset~\cite{Raven_19} contains a rich set of RPM tests.
Every RPM test consists of an incomplete 3×3 matrix of context panels, and eight candidate answer panels. %
The goal of solving an RPM test is to understand the row-wise underlining rule set, and then to decide which of the candidate panels is the most appropriate choice to complete the matrix. 
An example of RPM test can be seen in Supplementary Fig.~1.
The RAVEN dataset arranges the RPM tests in seven different constellations, namely center, 2x2 grid, 3x3 grid, left-right, up-down, out-in center, and out-in grid which are shown in Supplementary Fig.~1. 
The panels have a resolution of $r\times r=160\times 160$. 
The dataset provides 10,000 samples for every constellation, which are divided into six training folds, two validation folds, and two testing folds. 

The objects inside the panels have the following attributes: number, position, type, size, and color. 
RAVEN distinguishes between five different types (triangle, square, pentagon, hexagon, and circle), six sizes (enumerated from 1--6), and ten colors in the form of shadings (enumerated from 1--10).
The number of objects present in the panel varies from one to the maximum number of possible objects, which is determined by the constellation, e.g., the 2x2 grid contains maximally four objects.  
The position attribute describes the occupancy of the objects inside the panel.
Its range is constellation-dependent too; e.g., the 2x2 grid has 15 different position constellations. 
Nine panels are arranged to a 3x3 matrix such that one out of the following four rules applies to each attribute in a row-wise manner.  
The four types of rules can be summarized as: 
\begin{itemize}
    \item \texttt{Constant}: The attribute value does not change per row. 
    \item \texttt{Progression}: The attribute value monotonically increases or decreases in a row by a value of 1 or 2. 
    \item \texttt{Arithmetic}: The attribute values of the first two panels are either added or subtracted, yielding the attribute value of the third panel in the row. 
    \item \texttt{Distribute three}: This rule involves the fact that three different values of an attribute appear in the three panels of every row (with distinct permutations of the values in different rows). The same holds with respect to the columns.
\end{itemize}
Supplementary Fig.~1 shows an example for each rule governing the position attribute or the number attribute.

The answer choices in the RAVEN dataset are generated in a way such that only one randomly chosen attribute value differs from the correct answer. 
Consequently, by exploiting this shortcut
solution, the correct answer can be found by simply considering the mode of attribute values in the answer set without looking at the context panels, which is considered as unfair~\cite{I-Raven,MRNet_CVPR2021}. 
To this end, the impartial RAVEN (I-RAVEN~\cite{I-Raven}) provides an alternative answer set, which is generated with an attribute bisection tree ensuring that the modifications of attribute values are well balanced without any detectable pattern. 
\cc{It is therefore recommended to use the unbiased I-RAVEN when testing RPM reasoning models~\cite{RPM_Survey2022,Mitchel_Survey_2021}}.

\subsection*{\cc{PMF computation for object attributes and panel attributes}}
{\cc{
Every panel is represented with the attribute PMFs describing the distribution of the attribute values inside the panel.
First, the object PMFs are determined using marginalization with a consecutive softmax activation.
The marginalization computes the sum of non-negative cosine similarities between the query and each valid attribute value combination. 
For example, for the attribute type with value $j$ at location $k$, we determine the sum  by
\begin{align}
    \mathbf{v}'^{(k)}_{type}[j] = \sum_{\substack{s \in\lbrace 1,...,6 \rbrace \\ c\in\lbrace 1,...,10 \rbrace}} \mathrm{ReLU}\left(\mathrm{sim}(\mathbf{q},\mathbf{l}_k \odot \mathbf{t}_j \odot \mathbf{s}_s \odot \mathbf{c}_c  )\right), 
\end{align}
where $\mathrm{ReLU}(\cdot)$ is the rectified linear unit activation. 
Similarly, the sum is computed for the object PMF of attribute color and size. 
For marginalizing for object existence ($\mathbf{v}'^k_{exist}$[0]), we sum over all the attribute value combinations, whereas the value of no existence is given by the cosine similarity between the query and the vector ($\mathbf{f}_k$) which indicates that no object is present at position $k$: 
\begin{align}
    \mathbf{v}'^{(k)}_{exist}[0] &= \sum_{\substack{t \in\lbrace 1,...,5 \rbrace \\ s \in\lbrace 1,...,6 \rbrace \\ c\in\lbrace 1,...,10 \rbrace}} \mathrm{sim}(\mathbf{q}, \mathbf{l}_k \odot \mathbf{t}_t \odot \mathbf{s}_s \odot \mathbf{c}_c ) \\
    \mathbf{v}'^{(k)}_{exist}[1] &= \mathrm{sim}(\mathbf{q}, \mathbf{f}_k ). 
\end{align}
In this case, we marginalize over all cosine similarities (i.e., no ReLU activation is applied) as the $\mathbf{v}'^{(k)}_{exist}[1]$ would be 0 with high probability at the beginning of training, rendering the end-to-end training infeasible. 
Finally, for generating a valid object PMF which sums up to one, we apply a scaled softmax non-linearity on the sum vector. 
More precisely, for each attribute $a$ as either type, color, size, or exist, and the panel position $k$, we compute
\begin{align}
   \mathbf{v}^{(k)}_{a} &= \mathrm{softmax}(s_m \cdot \mathbf{v}'^{(k)}_{a}), 
\end{align}
where $s_m$ is a trainable, inverse softmax temperature.
}}

Next, the PMFs of the different objects are combined to five PMFs representing the attributes of the panel. 
The constellation is known to the reasoning backend; hence, the dimensions of the PMFs that depend on the constellations (i.e., position, number) are known, too. 
The position PMF represents the probability of object occupancy inside a panel. 
An occupancy $p$ is described with the set $I_p$ containing the occupied positions, e.g., $I_{1} =\lbrace 1 \rbrace$ represents the case where only the first object is occupied, and $I_{511}=\lbrace 1,2,3,...,9 \rbrace$ the case where all objects are occupied in a 3x3 grid. 
The position PMF is derived by
\begin{align}
    \mathbf{p}_{pos}[j] = \prod_{k\in I_{j}}\mathbf{v}^{(k)}_{exist}[0]\cc{\prod_{k'\in \{1,...,9\} \setminus I_{j}}\mathbf{v}^{(k')}_{exist}[1]}. 
\end{align}
For the attribute number, the PMF is derived from the position PMF with
\begin{align}
    \mathbf{p}_{num}[j] =\sum_{\substack{k=1 \\s.t. \, j=|I_{k}|}}^{n_{pos}} \mathbf{p}_{pos}[k], 
\end{align}
where $|I_{k}|$ represents the number of occupied positions in the set $I_{k}$. 
For the attributes type, size, and color, the PMFs are determined by combining the position PMF with the corresponding attribute PMFs. 
For example, the PMF for the attribute type is determined by \begin{align}
    \mathbf{p}_{type}[j]=\sum_{i=1}^{n_{pos}}\mathbf{p}_{pos}[i] \prod_{k \in I_i} \mathbf{v}_{type}^{(k)}[j]. 
\end{align}
In some RPM tests, the values of some attributes inside a panel can be different, e.g., the types are different. 
We represent this case with an inconsistency state for the attributes type, size, and color, by extending the PMF with an additional probability, e.g., for the attribute type
\begin{align}
    \mathbf{p}_{type}[n_{type}+1] = 1-\sum_{j=1}^{n_{type}}\mathbf{p}_{type}[j]. 
\end{align}

\subsection*{PMF transformation of discrete and continuous attributes to VSA}

The PMF of a discrete attribute is represented with a vector space which is spanned with unrelated basis vectors $B:=\lbrace \mathbf{b}_i \rbrace_{i=1}^n$. 
Each basis vector $\mathbf{b}_i \in \Binary^d$ is a $d$-dimensional $\kappa$-sparse binary vector, where the vector is divided in $\kappa$ blocks each containing one non-zero element whose index drawn from a uniform distribution. 
For representing the PMF of a continuous attribute, we use a vector space which is spanned with a basis taken from the fractional power encoding~\cite{PlateHolographic2003}. 
Building the basis of the fractional power encoding begins with randomly initializing one single unitary basis vector $\mathbf{e}\in \Binary^d$. 
The basis vector corresponding to any arbitrary attribute value $v$ is defined by exponentiation of the basis vector $\mathbf{e}$ using the value as the exponent.
For example, the basis vector corresponding to the value ``3'' is $\mathbf{e}^{3}=\mathbf{e}\odot\mathbf{e}\odot\mathbf{e}$. 
For representing real values $v\in\mathbb{R}$, the corresponding basis vector can be computed in the block-wise Fourier domain, where the final basis vectors can contain more than $\kappa$ non-zero elements~\cite{frady2021vfa}.
In RPMs, however, we exclusively encounter attributes with the integer values, e.g., the size attribute of an object is enumerated from 1 to $n=6$. 
Thus, the underlying codebook is $\kappa$-sparse, of finite size, and defined as $B:= \lbrace \mathbf{e}^i \rbrace_{i=1}^{n}$. 
Using this codebook, a PMF of a given continuous attribute is then transformed to the VSA representation using the weighed superposition defined in equation~\eqref{eq:block_superpos}. 

\subsection*{End-to-end training}
In the following, we give a detailed description of the end-to-end training of \name. 
We are given a training RPM task containing the panels $\mathbf{X}$ (8 context and 8 answer panels), the panel index of the ground-truth answer $y_{task}$, and the ground-truth rule $y_{rule}$. 
Here, we describe the updates of the frontend trainable parameters $\theta$ for one attribute, the generalization to all the rules is straightforward. 
The operation of \name results in the PMF estimation of the rule $\mathbf{u}$, the sampled rule $\tilde u$, and the scores $s$ derived from the negative JSD. 
Based on the scores, we compute the categorical cross-entropy loss $l(\mathbf{X},y_{task})$. 
Similar to PrAE~\cite{PrAE_CVPR21}, we update the trainable parameters $\theta$ using the following gradient update
\begin{align}\label{eq:reinforce}
\theta \leftarrow \theta - \beta \left( \nabla_{\theta} l(\mathbf{X},y_{task}) + l(\mathbf{X},y_{task})\nabla_{\theta} \mathrm{log}({\mathbf{u}[\tilde{u}}]) - \nabla_{\theta} \mathrm{log}(\mathbf{u}[y_{rule}]) \right), 
\end{align}
where $\beta$ is the learning rate. 
The first update term minimizes the cross-entropy loss, where the second term, based on REINFORCE, operates on the sampled action log probability. 
Finally, the third auxiliary term operates on the log probability of the ground-truth rule in order to improve the rule detection.

\subsection*{Experimental Setup}
We evaluate different methods on the RAVEN~\cite{Raven_19} and the I-RAVEN~\cite{I-Raven} datasets. 
Our \name is exclusively trained on the training data from RAVEN while being tested on both RAVEN and I-RAVEN.
I-RAVEN provides the unbiased answer panels, while the constellations and the context matrices stay the same as in RAVEN.
In our experiments, in the \name frontend we set the dimension of the bipolar vectors to $d$=512, while in the \name backend we set the dimension of the binary sparse block codes to $d$=1024 and $\kappa$=4.
%

We train a separate \name, consisting of the frontend (i.e., a trainable ResNet-18 with the frozen $\mathbf{W}$) and the backend, per constellation.
Motivated by~\cite{RPM_ImageNet_pretraining_2020}, the ResNet-18 was pre-trained on the ILSVRC2012 ImageNet-1k dataset.
{\ccc{We also adapted ResNet18's first convolutional block by reducing its stride from 2 to 1 and removing the maxpooling, which improved the overall accuracy by 2.9\% and 2\% on RAVEN and I-RAVEN, respectively.}}
The training was performed for 150 epochs using the Adam optimizer with a weight decay of 1e-4 and a constant learning rate of 9.5e-5. 
All the training hyperparameters are determined based on the end-to-end reasoning performance on the validation set of RAVEN.
We searched through possible batch sizes $\lbrace 4, 8, 16, 32 \rbrace$. %
For the training on the majority of the constellations the hyperparameter search yielded an optimal batchsize of 16. 
As an only exception, the 3x3 grid constellation had to be trained with batchsize of 8 due to the large space requirements.

The models are implemented in PyTorch (version 1.4.0) and trained and validated on a Linux using an NVIDIA Tesla V100 GPU with 32\,GB memory.
We repeat all experiments five times with a different random seed and report the average results and standard deviation to account for training variability.

\section*{References}
\bibliographystyle{naturemag}
\bibliography{bibliography}

\section*{Acknowledgements}
This work is supported by the Swiss National Science foundation (SNF), grant 200800. The authors would like to thank Salmane El Messoussi for helping with the generalization experiments, Ross W Gayler for insightful comments that contributed to the final shape of the manuscript, and Linda Rudin for the careful proofreading. We would also like to thank Alexander Gray, Lior Horesh, Kenneth Clarkson, Ismail Yunus Akhalwaya, Maxence Ernoult for fruitful discussions, and Chid Apte and Robert Haas for managerial support.

\clearpage

\FloatBarrier 

\begin{table*}[h]
\centering
\caption{\cc{Reasoning accuracy (\%) and CPU compute time (min) when solving 2000 examples on the RAVEN test sets using the ground-truth panel attributes. Experiments where conducted with Intel Xeon E5-2640 cores running at 2.4\,GHz. \name backend was configured with vector dimension $d$=1024 and $\kappa$=4 blocks.}}\label{tab:results-reasoning}
{\cc{
\begin{tabular}{lrrrrrr}
\toprule
& \multicolumn{3}{c}{Accuracy (\%)} & \multicolumn{3}{c}{\begin{tabular}[c]{@{}c@{}}CPU compute\\ time (min)\end{tabular}} \\
 \cmidrule(l){2-4} \cmidrule(l){5-7}
Method & \multicolumn{1}{c}{2x2} & \multicolumn{1}{c}{3x3} & \multicolumn{1}{c}{O-IG}& \multicolumn{1}{c}{2x2} & \multicolumn{1}{c}{3x3}& \multicolumn{1}{c}{O-IG}   \\ \cmidrule(r){1-1} \cmidrule(l){2-4} \cmidrule(l){5-7}
PrAE~\cite{PrAE_CVPR21}  &   94.67 & 94.21 & 95.68 &1.0     &  648.1   & 1.4      \\ 
Our unrestricted PrAE  &      98.82 &  97.50 &99.22& 1.1       &   15,408.5   & 2.2  \\
\name backend          &     99.19 &  96.89 &  99.55&   12.6     &   63.2  & 18.5         \\  
\bottomrule
\end{tabular}%
}}
\end{table*}

\begin{table*}[!ht]
\centering
\caption{End-to-end model accuracy (\%) on the RAVEN test set. \ccc{The upper part of the table shows the results reported in the literature, while in the lower part, we report the average accuracy $\pm$ the standard deviation over five runs with different seeds for our \name and the reproduced baselines.
The \name was either trained end-to-end (see equation~\eqref{eq:reinforce}) or with the auxiliary visual attribute labels (see Supplementary equation~(3)).}}\label{tab:results-overall-RAVEN}
{
\begin{tabular}{lllllllll}
\toprule
Method & \multicolumn{1}{c}{Avg} & \multicolumn{1}{c}{Center} & \multicolumn{1}{c}{2x2 grid} & \multicolumn{1}{c}{3x3 grid} & \multicolumn{1}{c}{L-R} & \multicolumn{1}{c}{U-D} & \multicolumn{1}{c}{O-IC} & \multicolumn{1}{c}{O-IG}  \\ \cmidrule(r){1-1} \cmidrule(r){2-9}
WReN~\cite{WReN}                &  14.7       & 13.1   & 28.6 & 28.3  & 7.5   & 6.3   & 8.4   & 10.6        \\
ResNet~\cite{Raven_19}                &  53.4       & 52.8  & 41.8 & 44.3    & 58.8   & 60.2   & 63.2  & 53.1       \\
ResNet+DRT~\cite{Raven_19}                &  59.6       & 58.1   & 46.5   & 50.4   & 65.8   & 67.1   & 69.1       & 60.1      \\
\ccc{Shah \textit{et al.}~\cite{nesy2022_knowledge}}                &  \ccc{67.5}       & \ccc{94.6}   & \ccc{53.1}   & \ccc{33.9}   & \ccc{85.0}   & \ccc{89.1}   & \ccc{89.8}       & \ccc{31.9}      \\
LEN~\cite{zheng2019abstract}                &  78.3       & 82.3   & 58.5 & 64.3    & 87.0  & 85.5   & 88.9  & 81.9       \\
CoPINet~\cite{CoPINET_19}         &  91.4      & 95.1   & 77.5  & 78.9 & 99.1  & 99.7 & 98.5     & 91.4   \\
DCNet~\cite{zhuo2021dcnet}         & 93.6 & 97.8 & 81.7 &  86.65 &  99.8 & 99.8& 99.0 & 91.5  \\
\cmidrule(r){1-9}
{PrAE~\cite{PrAE_CVPR21}}                &  60.3$^{\pm 2.7 }$ &  70.6$^{\pm 5.3 }$ & 83.5$^{\pm 2.8}$&30.5$^{\pm 1.7}$ & 88.5$^{\pm 4.2 }$ & 89.2$^{\pm 4.1 }$ & 37.3$^{\pm 3.1 }$ & 22.7$^{\pm 2.2 }$\\
{MRNet~\cite{MRNet_CVPR2021}}   & 74.7$^{\pm 3.3 }$ &  96.2$^{\pm 4.3 }$ & 49.1$^{\pm 4.7}$&45.9$^{\pm 6.0}$ & 93.7$^{\pm 2.1 }$ & 94.2$^{\pm 2.1 }$ & 92.5$^{\pm 1.1 }$ & 51.3$^{\pm 11.5 }$ \\
\ccc{SCL~\cite{wu2020scl}}   & \ccc{87.2$^{\pm 0.9 }$} &  \ccc{99.9$^{\pm 0.0 }$} & \ccc{76.6$^{\pm 0.8}$}& \ccc{63.5$^{\pm 4.7}$} & \ccc{96.8$^{\pm 5.6 }$} & \ccc{98.4$^{\pm 3.0 }$} & \ccc{96.5$^{\pm 3.6 }$} & \ccc{78.5$^{\pm 2.5}$} \\
\ccc{\name (end-to-end tr.)}   & \ccc{87.7$^{\pm 0.5 }$} &  \ccc{99.7$^{\pm 0.4 }$} & \ccc{93.5$^{\pm 2.9}$}&\ccc{57.1$^{\pm 3.3}$} & \ccc{99.8$^{\pm 0.1 }$} & \ccc{99.7$^{\pm 0.2 }$} & \ccc{98.6$^{\pm 1.6 }$} & \ccc{65.4$^{\pm 0.6 }$} \\
\ccc{\name (attribute label tr.)} & \ccc{98.5$^{\pm 0.1 }$} & \ccc{100$^{\pm 0.0 }$} & \ccc{99.4$^{\pm 0.0}$}&\ccc{ 96.3$^{\pm 0.6}$} & \ccc{100$^{\pm 0.0 }$} & \ccc{100$^{\pm 0.0 }$} & \ccc{100$^{\pm 0.0 }$} & \ccc{93.9$^{\pm 0.0 }$} \\
\cmidrule(r){1-9}
Human~\cite{Raven_19}                & 84.4      &  95.5 &  81.8 & 79.6   & 86.4   &  81.8 & 86.4       &  81.8     \\
\bottomrule
\end{tabular}%
  }
\begin{tablenotes}\footnotesize
\item[*]
\end{tablenotes}
\end{table*}

\begin{table*}[!ht]
\centering
\caption{End-to-end model accuracy (\%) on the I-RAVEN test set. \ccc{The upper part of the table shows the results reported in the literature, while in the lower part, we report the average accuracy $\pm$ the standard deviation over five runs with different seeds for our \name and the reproduced baselines.
The \name was either trained end-to-end (see equation~\eqref{eq:reinforce}) or with the auxiliary visual attribute labels (see Supplementary equation~(3)).}}\label{tab:results-overall-IRAVEN}
{
\begin{tabular}{lllllllll}
\toprule
Method &  \multicolumn{1}{c}{Avg} & \multicolumn{1}{c}{Center} & \multicolumn{1}{c}{2x2 grid} & \multicolumn{1}{c}{3x3 grid} & \multicolumn{1}{c}{L-R} & \multicolumn{1}{c}{U-D} & \multicolumn{1}{c}{O-IC} & \multicolumn{1}{c}{O-IG}  \\ \cmidrule(r){1-1} \cmidrule(r){2-9}
WReN~\cite{WReN}                &  23.8       &  29.4   &  26.8 &  23.5  & 21.9   &  21.4   &  22.5   &  21.5        \\
ResNet~\cite{Raven_19}                &  40.3       &  44.7   &  29.3 &  27.9    &  51.2   &  47.4   &  46.2  &  35.8       \\
ResNet+DRT~\cite{Raven_19}                &  40.4       &  46.5   &  28.8   &  27.3   &  50.1   &  49.8   &  46.0       &  34.2      \\

SRAN~\cite{I-Raven}                &   60.8       &  78.2   &  50.1 &  42.4    &  70.1   & 70.3   &  68.2  &  46.3       \\
LEN~\cite{zheng2019abstract}                &   41.4       &  56.4   &  31.7 &  29.7    & 44.2   & 44.2   & 52.1  &  31.7       \\
CoPINet~\cite{CoPINET_19}         &   46.1       &  54.4   &  36.8   &  31.9   & 51.9   &  52.5   &  52.2       &  42.8      \\
DCNet~\cite{zhuo2021dcnet}         &  49.36 &  57.8 &  34.1 &  35.5 &   58.5 & 60.0 &  57.0 &  42.9    \\
\cmidrule(r){1-9}
{PrAE~\cite{PrAE_CVPR21}}                &  71.1$^{\pm 2.1 }$ &  83.8$^{\pm 3.4 }$ & 82.9$^{\pm 3.3}$&47.4$^{\pm 3.2}$ & 94.8$^{\pm 2.1 }$ & 94.8$^{\pm 2.1 }$ & 56.6$^{\pm 3.0 }$ & 37.4$^{\pm 1.7 }$ 
\\
{MRNet~\cite{MRNet_CVPR2021}}   & 75.0$^{\pm 1.4 }$ &  96.8$^{\pm 3.7 }$ & 45.6$^{\pm 3.3}$&39.6$^{\pm 1.8}$ & 95.7$^{\pm 1.5 }$ & 95.9$^{\pm 1.8 }$ & 95.6$^{\pm 1.5 }$ & 55.5$^{\pm 4.7 }$ 
\\
\ccc{SCL~\cite{wu2020scl}}  & \ccc{84.3$^{\pm 1.1 }$} &  \ccc{99.9$^{\pm 0.0 }$} & \ccc{68.9$^{\pm1.9}$} & \ccc{43.0$^{\pm 6.2}$} & \ccc{98.5$^{\pm 2.9 }$} & \ccc{99.1$^{\pm 1.5}$} & \ccc{97.7$^{\pm 1.3 }$} & \ccc{82.6$^{\pm 2.5 }$} \\
\ccc{\name (end-to-end tr.)} & \ccc{88.1$^{\pm 0.4 }$} &  \ccc{99.8$^{\pm 0.2 }$} & \ccc{96.2$^{\pm 0.8}$}& \ccc{54.3$^{\pm 3.2}$} & \ccc{100$^{\pm 0.1 }$} & \ccc{99.9$^{\pm 0.1 }$} & \ccc{99.6$^{\pm 0.5 }$} & \ccc{67.1$^{\pm 0.4 }$} \\
\ccc{\name (attribute label tr.)} & \ccc{99.0$^{\pm 0.3 }$} & \ccc{100$^{\pm 0.0 }$} & \ccc{99.5$^{\pm 0.0}$}& \ccc{97.1$^{\pm 1.8}$} & \ccc{100$^{\pm 0.0 }$} & \ccc{100$^{\pm 0.0 }$} &\ccc{100$^{\pm 0.0 }$} & \ccc{96.4$^{\pm 0.0 }$} \\
\bottomrule
\end{tabular}%
}
\begin{tablenotes}\footnotesize
\item[*]
\end{tablenotes}
\end{table*}

\renewcommand{\figurename}{Supplementary Figure}
\renewcommand{\tablename}{Supplementary Table}
\renewcommand\refname{\textbf{Supplementary References}}
\setcounter{figure}{0}
\setcounter{equation}{0}
\setcounter{table}{0}

\clearpage
\section*{Supplementary Figures}
\subsection*{Supplementary Figure 1: Details on the RAVEN dataset.}
\begin{figure}[hb]
    \centering
    \fontsize{7}{10}
    \includegraphics[width=0.95\textwidth]{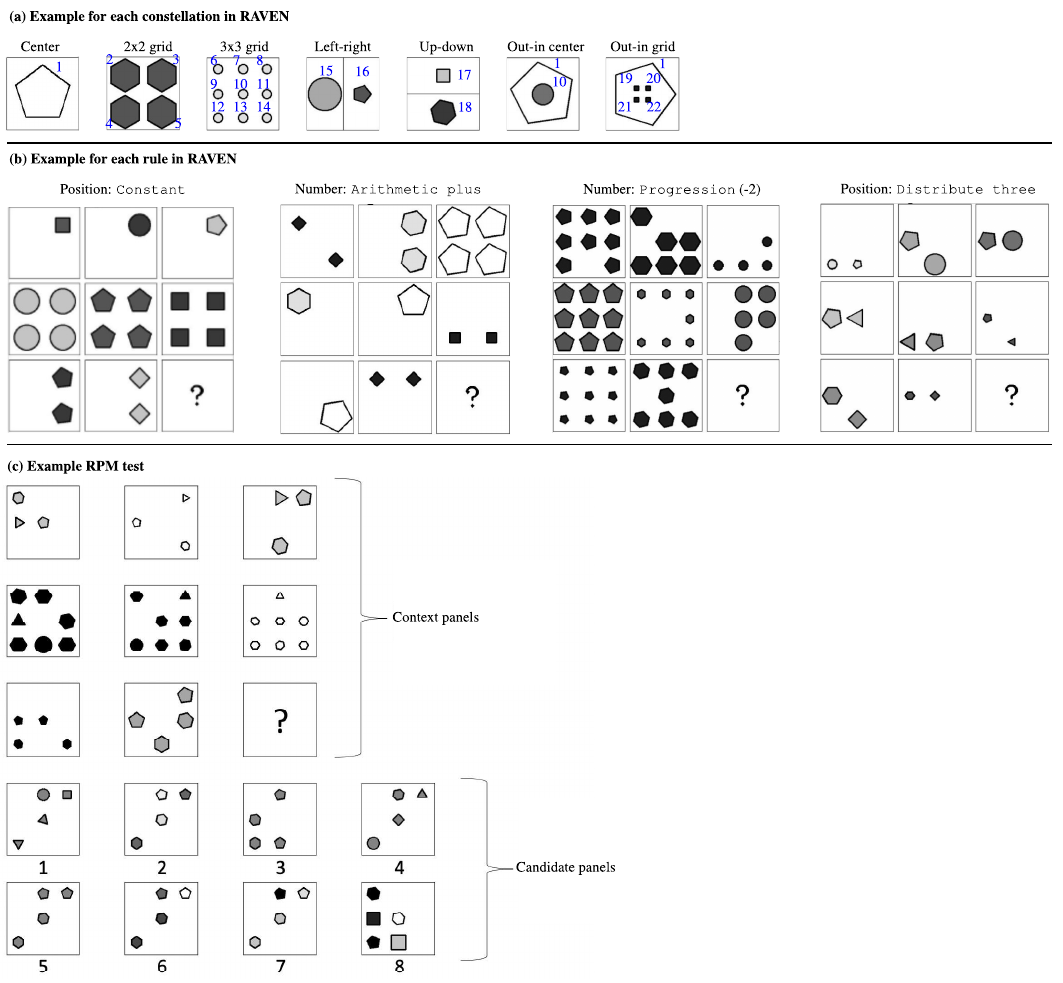}
    \caption{
    \textbf{(a)} Examples of the seven constellations in the RAVEN dataset. We enumerate 22 unique positions (in blue) across all seven constellations. Moreover, we merge overlapping positions with the same proportions across constellations, which are 1) the object in center, the outer object in out-in center, and the outer object in out-in grid (enumerated with ``1''); and 2)  the middle object in 3x3 grid and the inner object in out-in center (enumerated with ``10'').
    \textbf{(b)} Examples for the four types of rules in RAVEN. In these examples, the rules are applied on the position attribute, or number attributes. A separate rule is applied per attribute, the displayed attribute and rule is just one of them.
    \textbf{(c)} An example of RPM test from the RAVEN dataset using the 3x3 grid constellation. There are eight context panels and eight answer panels. In this example, the number of objects stays constant per row. Moreover, the size values (small, medium, large) of the objects are distributed per row. Even though the shapes do not agree within a panel, they stay constant per row. The arithmetic minus rule is applied on the attribute color. Combining the detected rule leads to the correct answer panel 5.}
    \label{fig:rpm}
\end{figure}

\clearpage

\clearpage

\section*{Supplementary Notes}
\subsection*{Supplementary Note 1: Neural network representation learning over VSA and its generalization}
In this Supplementary Note, we present further investigations into the \name frontend for the visual perception. 
In the first subsection (\textbf{a}), we explain the direct supervised training of the frontend in the presence of the visual ground-truth attribute labels using a novel additive cross-entropy loss.
We also analyze the object classification accuracy and compare it with other loss functions and perceptual methods. 
In the next two subsections (\textbf{b} and \textbf{c}), we study the generalizability of the \name frontend in isolation for unseen attribute-value combinations (\textbf{b}) and unseen combinations of multiple objects (\textbf{c}).

\subsubsection*{a. Supervised training with additive cross-entropy loss and comparisons}
We consider a supervised training setup in which the visual ground-truth attribute labels for all objects are provided.
Therefore, the frontend can be trained standalone.
We mutually train a universal \name frontend on all training constellations by enumerating all possible positions and merging the identical positions across constellations (see Supplementary Fig.~1). 
For an image panel $\mathbf{X}$, containing $k$ objects, with $k$ target indices $Y:=\lbrace y_i \rbrace_{i=1}^k $, the trainable parameters $\theta$ of ResNet-18 are optimized to maximize the similarity between its output query $\mathbf{q}=f_{\theta}(\mathbf{X})$ and the bundled vector $\mathbf{w}_{y_1}\oplus ... \oplus \mathbf{w}_{y_k}$. 
The dictionary matrix $\mathbf{W}$ stays fixed during training. 
As noted, each vector in $\mathbf{W}$ is computed by multiplicative binding of the codebooks, so we call this $\mathbf{W}$ encoding multiplicative binding.
Due to the similarity-preserving property of the bundling operation, maximizing the similarity between the query vector and the bundled vector is equivalent to maximizing the similarity between the query vector and each object vector:
\begin{align}\label{eq:superpos_nn}
    \theta ^* = &\operatornamewithlimits{argmax}_{\theta} \mathrm{sim} \left( f_{\theta}(\mathbf{X}), \mathbf{w}_{y_1}\oplus ... \oplus\mathbf{w}_{y_k}\right) 
    \\ 
    \approx & \operatornamewithlimits{argmax}_{\theta} \mathrm{sim} \left( f_{\theta}(\mathbf{X}), \mathbf{w}_{y_1}\right)+...+\mathrm{sim} \left( f_{\theta}(\mathbf{X}), \mathbf{w}_{y_k}\right).
\end{align}

We propose to optimize equation~\eqref{eq:superpos_nn} utilizing a novel additive cross-entropy loss, defined as 
\begin{align} \label{eq:acel}
     \mathcal{L}\left(\mathbf{X},Y,\theta \right) := -\mathrm{log} \frac{e^{s_l\cdot \left(\mathrm{sim}(f_{\theta}(\mathbf{X}),\mathbf{w}_{y_1})+...+\mathrm{sim}(f_{\theta}(\mathbf{X}),\mathbf{w}_{y_k}))\right)}}{e^{s_l\cdot \left(\mathrm{sim}(f_{\theta}(\mathbf{X}),\mathbf{w}_{y_1})+...+\mathrm{sim}(f_{\theta}(\mathbf{X}),\mathbf{w}_{y_k}))\right)} + \sum_{i=1}^m e^{s_l\cdot \left(\mathrm{sim}(f_{\theta}(\mathbf{X}),\mathbf{w}_{y_i})\right)} }, 
\end{align}
where $s_l$ is an inverse softmax temperature. 
The loss is optimized using the batched stochastic gradient descent by exclusively updating the parameters $\theta$ while freezing $\mathbf{W}$.
As the cosine similarity is bound between -1 and +1 and the softmax function embedded in the cross-entropy loss is scale sensitive, we scale the logit vector with a scalar $s_l$, serving as an inverse softmax temperature for improved training. 

As an alternative loss function, the \name frontend can be trained with a randomized cross-entropy loss, which focuses on optimizing of the similarity between the query and one randomly picked object vector $\mathbf{w}_{j_i,\, i\in\{y_1, ..., y_k\}}$ at a time.    
We compute the $m$-dimensional logit vector $\mathbf{z}=\mathbf{W}\mathbf{q}$, pick one of the target indices at random, and compute the cross-entropy loss based on the scaled logit vector and the randomly picked target index.
By repeating the optimization for multiple epochs, the randomized cross-entropy loss guides $f_{\theta}$ to generate a composite vector that resembles the bundling of all object vectors in the panel.

During inference, ResNet-18 generates a query vector that can be decomposed into constituent object vectors.
The decomposition performs a matrix-vector multiplication between the normalized dictionary matrix $\mathbf{W}$ and the normalized query vector, $\mathbf{q}$, to obtain the cosine similarity scores $\mathbf{z}$.
%
The similarity scores are passed through a thresholded detection function $g_{\tau}$, which returns the indices of the score vector whose similarity exceeds a threshold. 
The optimal threshold $\tau:=0.23$ is determined by cross-validation and is identical across all constellations.
Since the structure of the dictionary matrix is known, we can infer the labels for the attributes, namely position, color, size, and type, from the detected indices. 

In the following, we assess the performance of the \name frontend by evaluating the panel accuracy when predicting the attribute values of type, size, color, and position for each panel. 
A correct prediction is counted only if all attribute values of all objects in a panel are predicted correctly. 
We compare the perception accuracy of the \name frontend in different training configurations with the visual perception part of PrAE~\cite{PrAE_CVPR21}.
The visual perception part of PrAE consists of four separate LeNet-like architectures, which predict objectiveness, type, size, and color. 
Since the original PrAE was trained only on the 2x2 constellation, we also train the visual perception part of PrAE on each constellation individually.

For learning the parameters of our \name frontend and PrAE~\cite{PrAE_CVPR21}, we extract the 16 panels (eight context panels and eight answer panels) and use ground-truth attribute values provided by the dataset as meta-labels. 
We exclusively trained and tested the models on the RAVEN training and testing sets, respectively. 
Moreover, we also train our \name frontend on a partial training set containing only 6000 training samples (instead of full 42,000 samples) by taking training samples from the individual constellations based on a share that corresponds to their number of possible locations, e.g., 3x3 grid provides 9$\times$ more training samples than the center. 
The trainable parameters are trained using batched stochastic gradient descent (SGD) with a weight decay of $10^{-4}$1e-4 and a momentum of 0.9.

Supplementary Table~\ref{tab:results-perception-new} compares the panel accuracy of these different perception methods. 
Training the \name frontend on the full training set yields a highly accurate model that significantly outperforms the constellation-dependent PrAE models, where the additive cross-entropy loss results in 2.4\% higher accuracy compared to the random loss (99.76\% vs. 97.33\%).  
The additive cross-entropy loss notably outperforms the randomized cross-entropy loss in the constellations with many possible locations, e.g., in the 3x3 grid (98.61\% vs. 85.70\%) or the out-in grid (99.95\% vs. 97.30\%). 
Moreover, when training the perception only on the partial training set (i.e., $1/7$ of the full training set), the \name frontend accuracy is almost preserved with both the additive (99.76\% vs. 97.16\%) and the random cross-entropy loss (97.33\% vs. 96.78\%) while reducing the training set to the size of a single constellation (42,000 samples vs. 6000 samples). 
This showcases the sample efficiency of our approach.

Finally, we merge this instance of the \name frontend, which is trained on the complete training set with the additive cross-entropy loss, with the \name backend to solve the complete RPM tests. 
Tables~II and III show the performance in the last row. 
\name achieves the highest accuracy of 98.5\% and 99.0\% on RAVEN and I-RAVEN, respectively, thanks to its accurate perception. 

\begin{table*}[h]
\centering
\caption{Panel accuracy (\%) of the visual perception methods on the RAVEN test set. The methods are trained with the visual attribute labels. Avg denotes the average accuracy over all test constellations. L-R stands for left-right, U-D for up-down, O-IC for out-in center, and O-IG for out-in grid. }\label{tab:results-perception-new}
\begin{tabular}{lrrrllllllll}
\toprule
Method & \multicolumn{1}{c}{\begin{tabular}[c]{@{}c@{}}Training \\ Loss\end{tabular}}& \multicolumn{1}{c}{\begin{tabular}[c]{@{}c@{}}Training \\ Constellation(s)\end{tabular}} &\multicolumn{1}{c}{\begin{tabular}[c]{@{}c@{}}\# Training \\ Samples\end{tabular}} & \multicolumn{1}{c}{Avg} & \multicolumn{1}{c}{Center} & \multicolumn{1}{c}{2x2 grid} & \multicolumn{1}{c}{3x3 grid} & \multicolumn{1}{c}{L-R} & \multicolumn{1}{c}{U-D} & \multicolumn{1}{c}{O-IC} & \multicolumn{1}{c}{O-IG}  \\ \cmidrule(r){1-4} \cmidrule(r){5-12}

PrAE~\cite{PrAE_CVPR21} perception & Rand. CEL & Individual$^{\star}$ &42,000&   85.27       & 88.65   & 93.56   & 73.95   & 100.0   & 100.0   & 94.23       & 46.52
\\
\name frontend   & Add. CEL  & Full$^{\dagger}$  & 42,000 & 99.76 & 100 & 99.83 & 98.61 & 99.97 & 99.96 & 99.97 & 99.95\\
\name frontend   & Rand. CEL  & Full$^{\dagger}$  & 42,000 &  97.33 & 100 & 99.30 & 85.70 & 99.67 & 99.56 & 99.73 & 97.30 \\
\name frontend& Add. CEL  & Partial$^{\dagger}$ &6,000 & 97.16 & 98.84 & 99.26 & 86.23 & 99.66 & 99.55 & 99.75 & 96.85 \\
\name frontend& Rand. CEL      & Partial$^{\dagger}$ &6,000 &  96.78 & 99.83 & 99.18 & 84.95 & 99.63 & 99.52 & 99.74 & 94.57 \\ 
\bottomrule
\end{tabular}%
\begin{tablenotes}\footnotesize
\item[*] $^{\star}$ The training constellation is identical to the testing constellation, thus seven independent models were trained and tested.
\item[*] $^\dagger$ A universal model on all seven constellations was trained and tested.
\end{tablenotes}
\end{table*}

\subsubsection*{b. Generalizability of multiplicative binding to unseen combinations of attribute values}
In this part, we investigate the generalizability of the \name frontend to unseen attribute-value combinations. 
To that end, we consider the single object case for the 2x2 grid constellation. 
After choosing a set of values for each of the four attributes (position, color, size, and type), the training and test datasets for each of the six pairs of attributes are generated as follows.
For the pair of attributes $\{A_i, A_j\}_{i,j \in \{1,..,4\}}$ and their associated set of values $V_i$ and $V_j$, an object is considered during training if at least one of its values for $A_i$ or $A_j$ is in the prespecified sets $V_i$ and $V_j$. 
The test set only contains objects that do not satisfy the condition for $A_i$ nor $A_j$.

For instance, when we choose the first quadrant for position and the triangle for type as the pair of attribute-value of interest, the training set contains panels with single objects of all types placed on the first quadrant or triangles located on the remaining three quadrants.
In this case, the test set is composed of single objects of all types except triangles placed on all quadrants, excluding the first one. We note that in this setting, the considered objects can have any value for the rest of attributes that is color and size. The datasets for the example above are depicted in Supplementary Fig.~\ref{fig:attr_value_generalizability}.
Supplementary Table~\ref{tab:generalization_mult} lists the training and testing attribute combinations for all six pairs of attributes. 
\begin{figure}[!ht]
\centering
    \subfloat[Training set]{\includegraphics[width=.27\textwidth]{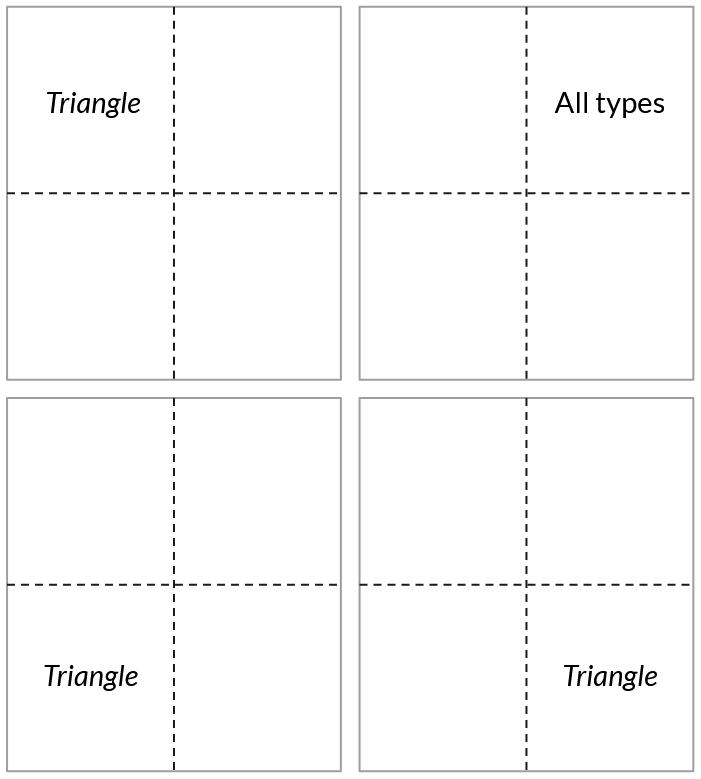}\label{fig:trainng_set}}
    \hspace{0.8cm}
    \subfloat[Test set]{\includegraphics[width=.27\textwidth]{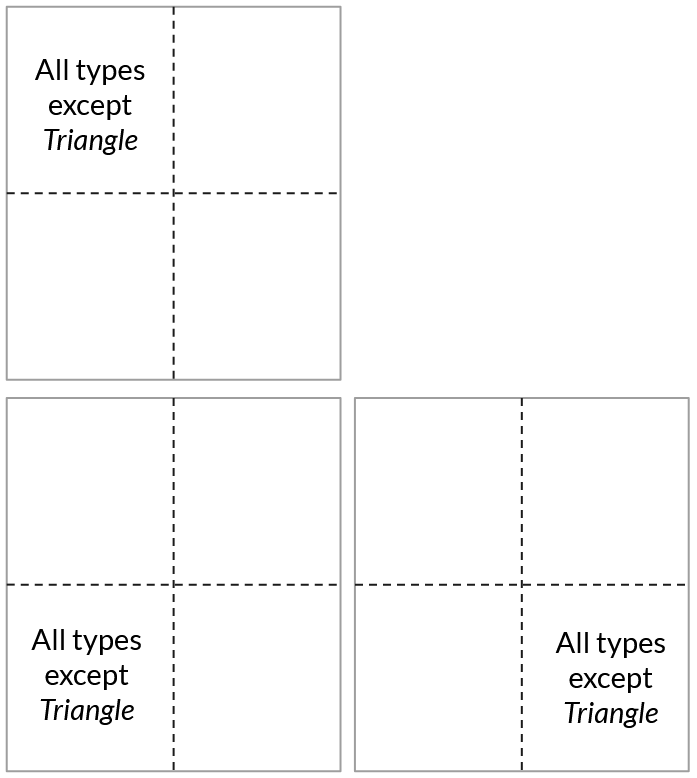}\label{fig:test_set}}
\caption{\cc{The types of panels considered in the training set (\textbf{a}) and test set (\textbf{b}) for the first type of generalization experiments in the following pair of attribute-value: the first quadrant for position, and triangle for type.}} \label{fig:attr_value_generalizability}
\end{figure}

The second type of generalization experiment involves determining two sets of values per attribute. For attribute $A_i$, the two sets are denoted $V_{i,1}$ and $V_{j,2}$. Accordingly, for the pair of attributes $\{A_i, A_j\}_{i,j \in \{1,..,4\}}$ and their respective four sets of values $V_{i,1}$, $V_{i,2}$, $V_{j,1}$, and $V_{j,2}$, the training set comprises objects with values for the two attributes in $V_{i,1}$ and $V_{j,1}$, or in $V_{i,2}$ and $V_{j,2}$. The test set is the remaining data points in the complement of the training set. 
This partitioning is inspired by the CLEVR dataset~\cite{CLEVR_CVPR2017} for compositional generalization experiments. 

In both investigated generalizability settings, the \name frontend \ccc{based on the multiplicative binding cannot provide the correct predictions for the test sets, resulting in 0\% test accuracy for all six attribute pairs (see the 4th column of Supplementary Table~\ref{tab:generalization_mult}).}
According to these results, this frontend instance does not show any sign of attribute-value generalizability. 
In fact, the frontend’s inability to generalize is an inherent property of the multiplicative binding of quasi-orthogonal vectors; each entry of the cosine similarity scores vector $\mathbf{z}$ corresponds to the similarity value with the embedding vector of a given combination of the considered four attributes. Through the additive cross-entropy loss, the model learns to maximize the entry corresponding to the target and minimize the rest. In the context of the generalization experiments, the set of components of the vector $\mathbf{z}$ corresponding to the training set and that of the test set are mutually exclusive. Therefore, when the dictionary matrix $\mathbf{W}$ contains quasi-orthogonal vectors, the model cannot be expected to perform accurate predictions for any unseen combination of the attributes.

After identifying this limitation in the attribute-value generalizability of the multiplicative binding encoding, we enhance this encoding by the addition (bundling) operation to add key-value bound pairs.
In this enhanced multiplicative-additive encoding, we describe an object $\mathbf{w}$ with attribute values $a,b,c,d$ for type, size, color, and position as 
\begin{align}
    \mathbf{w} = \left(\mathbf{r}_{\text{type}}\odot \mathbf{t}_a \right) \oplus \left(\mathbf{r}_{\text{size}}\odot \mathbf{s}_b\right) \oplus \left(\mathbf{r}_{\text{color}}\odot \mathbf{c}_c\right) \oplus \left(\mathbf{r}_{\text{pos}}\odot \mathbf{l}_d\right),  
\end{align}
where $\mathbf{r}_i \in \Bipolar^d$ are randomly initialized key vectors. 
The key vectors are bound with the corresponding value vectors ($\mathbf{t}_a$, $\mathbf{s}_b$, $\mathbf{c}_c$, and $\mathbf{l}_d$), yielding key-value pairs which are added (bundled) to represent all attributes of the object. 
The multiplicative-additive encoding allows the extraction of each attribute's value individually by unbinding the object representation ($\mathbf{w}$) with the key vector; the knowledge of other attribute values is not required.
Hence, this encoding explicitly disentangles the attribute values. 
This allows us to treat each attribute individually and, more importantly, formulate the attribute recognition as a regression problem where a relationship between values exists.
Concretely, we formulate the recognition of the color and the size as a regression problem using hyperspherical prototypes~\cite{mettes_hyperspherical}. 
Instead of dictating the target vector of every attribute value, hyperspherical regression only defines the target vectors of the minimum and maximum values. 
The intermediate values are uniformly distributed on the hypersphere in terms of cosine similarities. 
For example, the target vectors for the attribute color with 10 values are $\mathbf{c}_{1}=-\mathbf{x}$ and $\mathbf{c}_{10}=\mathbf{x}$, where $\mathbf{x}$ is a randomly initialized vector. 
The representation of an intermediate color value ($\mathbf{c}_{i}$) should then have a cosine similarity of 
\begin{equation}
\label{equ_hyperspherical}
\mathrm{cos}( \textbf{c}_{10}, \mathbf{c}_{i}) = 2 \cdot \frac{i - 1}{m_c -1} -1, 
\end{equation}
where $m_c=10$ is the number of possible color values. 
Finally, the visual perception module is trained by optimizing the mean-squared error for the hyperspherical prototypes (color and size) and the categorical cross-entropy loss for the position and type. 
%

Supplementary Table~\ref{tab:generalization_mult} compares the generalization capabilities of the \name frontend when using different encodings: the multiplicative versus the multiplicative-additive. 
Indeed, the multiplicative-additive encoding significantly improves the generalization in four attribute pairs compared to the pure multiplicative encoding: position-color (34.8\%), position-size (15.1\%), color-size (29.3\%), and color-type (72\%). 
There are still two attribute pairs of position-type and size-type that show 0\% generalization, which could be due to the spatial structure of the CNN's filters.

\begin{table}[]
\caption{\ccc{Accuracy of attribute-value generalization (\%) of the \name frontend in the 2x2 grid constellation containing $k$=1 object.
The training and test sets are chosen such that the attribute-value sets are disjoint. 
The NVSA frontend is trained with the ground-truth attribute labels by optimizing the loss in equation~\eqref{eq:acel} with SGD.
The NVSA frontend uses two different encodings: multiplicative binding of quasi-orthogonal vectors, and multiplicative-additive of hyperspherical. 
}}
\label{tab:generalization_mult}
\centering
{\ccc{
\begin{tabular}{lccrr}
\toprule
               & Training combinations                                                                     & Testing combinations                                                                               & \multicolumn{1}{c}{\begin{tabular}[c]{@{}c@{}}Multiplicative +\\ quasi-orthogonal\end{tabular}} & \multicolumn{1}{c}{\begin{tabular}[c]{@{}c@{}}Multiplicative-additive +\\ hyperspherical\end{tabular}} \\
\cmidrule(r){1-1}\cmidrule(r){2-2}\cmidrule(r){3-3}\cmidrule(r){4-4}\cmidrule(r){5-5}
Position-color & \begin{tabular}[c]{@{}c@{}}Position $\in$ \{0, 3\} OR\\ color $\in$ \{0, 3, 6, 8\}\end{tabular} & \begin{tabular}[c]{@{}c@{}}Position $\notin$ \{0, 3\} AND\\ color $\notin$ \{0, 3, 6, 8\}\end{tabular} & 0.0                                                                                      & 34.8                                                                                                   \\
\cmidrule(r){1-1}\cmidrule(r){2-2}\cmidrule(r){3-3}\cmidrule(r){4-4}\cmidrule(r){5-5}
Position-type  &   \begin{tabular}[c]{@{}c@{}}Position $\in$ \{0, 3\} \\OR type $\in$ \{0, 2\}\end{tabular} & \begin{tabular}[c]{@{}c@{}}Position $\notin$ \{0, 3\} \\AND type $\notin$ \{0, 2\}\end{tabular}    & 0.0                                                                                      & 0.0                                                                                                    \\
\cmidrule(r){1-1}\cmidrule(r){2-2}\cmidrule(r){3-3}\cmidrule(r){4-4}\cmidrule(r){5-5}
Position-size  &      \begin{tabular}[c]{@{}c@{}}Position $\in$ \{0, 3\} \\OR size $\in$ \{1, 5\}\end{tabular} & \begin{tabular}[c]{@{}c@{}}Position $\notin$ \{0, 3\} \\AND size $\notin$ \{1, 5\}\end{tabular}                                                                                                      & 0.0                                                                                      & 15.1                                                                                                   \\
\cmidrule(r){1-1}\cmidrule(r){2-2}\cmidrule(r){3-3}\cmidrule(r){4-4}\cmidrule(r){5-5}
Color-size     &      \begin{tabular}[c]{@{}c@{}}Color $\in$ \{0, 3, 6, 8\} \\OR size $\in$ \{1, 5\}\end{tabular} & \begin{tabular}[c]{@{}c@{}}Color $\notin$ \{0, 3, 6, 8\} \\AND size $\notin$ \{1, 5\}\end{tabular}                                                                                                    & 0.0                                                                                      & 29.3                                                                                                   \\
\cmidrule(r){1-1}\cmidrule(r){2-2}\cmidrule(r){3-3}\cmidrule(r){4-4}\cmidrule(r){5-5}
Color-type     &     \begin{tabular}[c]{@{}c@{}}Color $\in$ \{0, 3, 6, 8\} \\OR type $\in$ \{0, 2\}\end{tabular} &      \begin{tabular}[c]{@{}c@{}}Color $\notin$ \{0, 3, 6, 8\} \\AND type $\notin$ \{0, 2\}\end{tabular}  & 0.0                                                                                      & 72.0                                                                                                   \\
\cmidrule(r){1-1}\cmidrule(r){2-2}\cmidrule(r){3-3}\cmidrule(r){4-4}\cmidrule(r){5-5}
Size-type      &     \begin{tabular}[c]{@{}c@{}}Size $\in$ \{1, 5\} \\OR type $\in$ \{0, 2\}\end{tabular} &    \begin{tabular}[c]{@{}c@{}}Size $\notin$ \{1, 5\} \\AND type $\notin$ \{0, 2\}\end{tabular} & 0.0                                                                                      & 0.0                \\                                                                                   
\bottomrule
\end{tabular}
}}
\end{table}

\subsubsection*{c. Generalizability of multiplicative binding to unseen combinations of multiple objects}
{\cc{
In the previous subsection, \ccc{we observe that the \name frontend using the encoding of multiplicative-additive with hyperspherical prototypes can generalize to some unseen combinations of the attribute values in a single object, while the encoding with the multiplicative binding of the quasi-orthogonal vectors cannot.
Here, we further evaluate whether the multiplicative encoding can generalize to unseen combinations of multiple objects.}
We train the \name frontend (hereafter, we simply omit the repetitive multiplicative encoding term) in the 2x2 grid constellation where the training set contains as a basis all possible panels with exactly one object, which are 9600 panels when considering that we have 10 color, 6 size, 5 type, 8 angle, and 4 position attribute values. 
We provide two training settings k$_{\text{train}} \in \{1,2\}$, where k$_{\text{train}}$ is the number of available objects in the panel. 
In the training setting k$_{\text{train}}=1$, we only train the \name frontend using the basis training set with a single object, whereas in the k$_{\text{train}}=2$ setting we have augmented the basis training set by another 9,600 panels containing 2 objects. 
The validation sets are always constructed in the same way as the training sets.
In the testing, we consider the settings k$_{\text{test}} \in \{2,3,4\}$, where in each setting, the trained models are tested on 28,800 panels containing a fixed number of k$_{\text{test}}$ objects in the panel. 
See the first two rows in Supplementary Table~\ref{tab:results-superpos-generalization}.

The model parameters are trained using SGD with a weight decay of 1e-4 and a momentum of 0.9. The batchsize was set to 256, and we used the scaling factor $s_l=1$. Furthermore, we set the learning rate to 0.1 and decay by factor of 10 every 30 epochs. Moreover, the number of epochs is scaled such that all trained models have approximately the same number of updates. 
The optimal threshold $\tau$ is determined by cross-validation, where the selection criteria is $\tau = \argmax_{\hat{\tau}} v(\hat{\tau}) -4\hat{\tau}$, where $v(\tau)$ is the accuracy on the validation set using threshold $\tau$.
We included the regularization on the magnitude of $\tau$ since we generally predicted too few objects when k$_{\text{test}}$ was large.

Next, we construct a similar experiment in the 3x3 grid constellation, in which the basis of the training set contains all 21,600 possible single-object panels. 
Compared to the 2x2 grid, there are 9 position attributes instead of 4. 
We consider the training settings k$_{\text{train}} \in \{1,2,3,4\}$, where in k$_{\text{train}}= 1$ we only use the basis training set to train the \name frontend. 
In the settings where k$_{\text{train}} \geq 2$, we augment the training set which is used in the setting k$_{\text{train}}-1$ by 21,600 panels containing exactly k$_{\text{train}}$ objects. 
The validation sets are always constructed in the same way as the training sets in each setting. 
In the testing, we consider the settings k$_{\text{test}} \in \{2,3,4,5,6,7,8,9\}$, where in each setting, the trained models are tested on 64,800 panels containing a fixed number of k$_{\text{test}}$ objects in the panel.
The training hyperparameters are chosen as in the above experiment, except that the optimal threshold $\tau$ is determined using $\tau = \argmax_{\hat{\tau}} v(\hat{\tau}) -9\hat{\tau}$ as our selection criteria. 
Note that in both experiments, we omit to test the single object setting because every possible single object panel is already contained in the corresponding training set.
The results of the two experiment sets are summarized in Supplementary Table~\ref{tab:results-superpos-generalization}.

In the 2x2 grid constellation, we observe that after training in the k$_{\text{train}}=1$ setting, the \name frontend is already able to correctly predict the majority of the panels in the k$_{\text{test}} \in \{2,3\}$ settings, where it achieves 92.4\% and 68.1\% without even having seen an instance of multiple object panel in training. 
Nevertheless, there is a significant accuracy drop in the k$_{\text{test}}=4$, achieving 19.3\%. 
However, training with the 2 object combinations (i.e., k$_{\text{train}}=2$) achieves an average panel accuracy of 97.1\% in all testing settings.
Considering the out-of-distribution (OOD) testing cases, i.e., k$_{\text{test}} \in \{3,4\}$, an average panel accuracy of 97.3\% is achieved. 
This indicates that training with the simple cases of up to 2 objects in the panels is enough to generalize to panels that contain up to 4 objects.

In the 3x3 grid constellation, we observe similar trends in the k$_{\text{train}}=1$ setting, where it correctly predicts 70.6\% of the panels containing 2 objects. 
However, in the testing settings k$_{\text{test}} \in \{5,6,7,8,9\}$ the model is overwhelmed by the presence of too many objects at the same time. 
In k$_{\text{train}}=2$, it is able to obtain non-zero panel accuracy in all testing settings except the most complex one (k$_{\text{test}}=9$). 
This indicates that the model is able to generalize in more complex panels with up to 8 objects after having encountered the panels with at most 2 objects. 
We observe that increasing the number of objects during training results in a higher OOD average panel accuracy, where the accuracy is monotonically improving with respect to k$_{\text{train}}$.
Note that the OOD average is calculated over a more complex set, when we increase k$_{\text{train}}$. 
Finally, the \name frontend achieves an average panel accuracy of 86.3\% after training only on the panels which contain less than half of the maximal number of objects allowed in the 3x3 grid constellation.
}}

\begin{table}[h]
\caption{\cc{The \name frontend using multiplicative binding and its generalization to a growing number of unseen objects in the RAVEN panel. The frontend is trained with a fixed number of objects k$_{\text{train}}$ ranging from 1 to 2 in the 2x2 constellation, and then the test panel accuracy (\%) is reported for an unseen number of object combinations (k$_{\text{test}}$) ranging from 2 up to 4 objects. Similar experiments are done in the 3x3 constellation where k$_{\text{train}} \in \{1,2,3,4\}$ and k$_{\text{test}} \in \{2,3,4,...,9\}$. Avg denotes the average accuracy over all testing samples and OOD Avg denotes the average accuracy on testing samples with more than k$_{\text{train}}$ objects.} }\label{tab:results-superpos-generalization}
\centering
\begin{tabular}{cccccccccccccc}
\toprule
\multirow{2}{*}{\begin{tabular}[c]{@{}c@{}}Training\\ Constellation\end{tabular}} & \multirow{2}{*}{\begin{tabular}[c]{@{}c@{}}\# Training\\ Samples\end{tabular}} & \multirow{2}{*}{\# Epochs} & \multirow{2}{*}{$k_{\text{train}}$} & \multicolumn{8}{c}{$k_{\text{test}}$}                              & \multirow{2}{*}{Avg} & \multirow{2}{*}{\begin{tabular}[c]{@{}c@{}}OOD\\ Avg\end{tabular}} \\  \cmidrule(r){5-12}
&   &   &    & 2    & 3    & 4    & 5    & 6    & 7    & 8    & 9    &                      &                                                                    \\ \cmidrule(r){1-4} \cmidrule(r){5-14}
2x2 & 9600  & 200  & 1                       & 92.4 & 68.1 & 19.3 & -    & -    & -    & -    & -    & 59.9                 & 59.9                                                               \\
2x2                                                                               & 19,200                                                                        & 100                       & 2                       & 96.6 & 97.3 & 97.3 & -    & -    & -    & -    & -    & 97.1                 & 97.3                                                               \\ \cmidrule(r){1-14}
3x3                                                                               & 21,600                                                                        & 400                       & 1                       & 77.6 & 30.1 & 2.5  & 0.0  & 0.0  & 0.0  & 0.0  & 0.0  & 13.8                 & 13.8                                                               \\
3x3                                                                               & 43,200                                                                        & 200                       & 2                       & 89.2 & 83.4 & 60.9 & 30.7 & 10.9 & 2.8  & 0.6  & 0.0  & 34.8                 & 27.0                                                               \\
3x3                                                                               & 64,800                                                                        & 133                       & 3                       & 92.6 & 92.6 & 90.7 & 82.4 & 67.3 & 49.1 & 32.7 & 17.3 & 65.6                 & 56.5                                                               \\
3x3                                                                               & 86,400                                                                        & 100                       & 4                       & 89.3 & 89.9 & 90.9 & 91.8 & 91.4 & 88.4 & 81.2 & 67.8 & 86.3                 & 84.0                                                               \\ \bottomrule
\end{tabular}
\end{table}

\subsubsection*{d. Resolution issues in the RAVEN dataset}

In addition to its high perception accuracy, the \name frontend offers better transparency, allowing us to discover issues in the generative process in the RAVEN dataset.
Some objects in the inner part of the out-in grid constellation have a different size attribute value but the same image representation.
The problem occurs in cases where the size attribute differs by one value; hence, it can be attributed to an insufficient image resolution. 
This generation problem only concerns objects of type square and is observed in 42.15\% of the panels. 
For validating the perception accuracy, we solve this issue by merging classes with different sizes but same image representation. 

\clearpage

\subsection*{\cc{Supplementary Note 2: Visual Analogies}}
In this Supplementary Note, we demonstrate another use case in which the appropriate perceptual representations in the \name frontend can be used directly to solve higher-level reasoning tasks. 
Specifically, we show that the predicted perceptual representations at the output of ResNet-18 can be directly manipulated by the binding operations to solve visual analogy tasks ($A:B :: \alpha:\beta$). 
In the studied task, we consider a source domain that shares one relationship, or multiple relationships, between its two sets of objects ($A:B$), and a target domain that shares the same relationship(s) between its object sets ($\alpha:\beta$).
Binding the neural network representations obtained from the source domain allows to capture the relationship(s) solely from a single example, which can be applied to novel circumstances in the target domain by another application of the binding operation.

We generate a new RAVEN-like test dataset, in which a visual analogy problem consists of four panels arranged in a 2$\times$2 matrix with a missing panel in the bottom right, as shown in Supplementary Fig.~\ref{fig:one-to-one}. 
The first row constitutes the source domain where there is at least one relationship between the two panels (indicated by the blue arrow), and the second row constitutes the target domain that should establish the same relationship between one of its panels and the missing one.  
We demonstrate how the relationship can be captured from a single example in the source domain and how it can be applied beyond the example from which it was learned (i.e., in the novel circumstances of the target domain).
This can be done by solely applying consecutive binding operations at the vector outputs generated from the neural network: the first binding operation captures the relationship in the source domain, and the second one applies it to the target domain.

\subsubsection*{a. One-to-one relationship}
We describe how the relationship can be captured from the source domain and how it can be applied (i.e., transferred) to the target domain.
We explain it using the visual analogy example shown in Supplementary Fig.~\ref{fig:one-to-one}. 
We name this analogy problem \texttt{one\_to\_one} because there is only one relationship among the two objects in the source domain.
Using our \name frontend, ResNet-18 generates the VSA representations of $\mathbf{q}_A$, $\mathbf{q}_B$, and $\mathbf{q}_\alpha$ for the objects in panels $A$, $B$, and $\alpha$.
By manipulating these VSA representations, we aim to generate the VSA representation of the solution panel $\beta$.

To better explain the analogy, let us refer to the ground-truth VSA representations of the object in the panels.
We only refer to them for the sake of clarification; note that they are not used for solving analogies.
In our example shown in Supplementary Fig.~\ref{fig:one-to-one}, we have the following ground-truth VSA object representations:
\begin{align}
    \mathbf{o}_{A} = 
    \mathbf{l}_5 \odot \mathbf{t}_{\text{square}} \odot \mathbf{s}_2 \odot \mathbf{c}_3 \\
    \mathbf{o}_{B} = 
    \mathbf{l}_5 \odot \mathbf{t}_{\text{pentagon}} \odot \mathbf{s}_5 \odot \mathbf{c}_3 \label{eq:object_B}\\
    \mathbf{o}_{\alpha} = 
    \mathbf{l}_2 \odot \mathbf{t}_{\text{square}} \odot \mathbf{s}_2 \odot \mathbf{c}_3\\
    \mathbf{o}_{\beta}  = \mathbf{l}_2 \odot \mathbf{t}_{\text{pentagon}} \odot \mathbf{s}_5 \odot \mathbf{c}_3
\end{align}
In this example, $\mathbf{o}_A$ is related to $\mathbf{o}_B$ by changing its type from square to pentagon and increasing its size from 2 to 5.
Using the binding operation between the corresponding perceptual representations generated by ResNet-18 ($\mathbf{q}_{A}$ and $\mathbf{q}_{B}$) allows capturing this relationship in a VSA representation ($\mathbf{r}_{A:B}$) as a high-dimensional vector via:
\begin{align}
    \mathbf{r}_{A:B} = \mathbf{q}_{A} \odot \mathbf{q}_{B} \approx 
    \mathbf{o}_{A} \odot \mathbf{o}_{B} = \mathbf{t}_{\text{square}} \odot \mathbf{t}_{\text{pentagon}} \odot \mathbf{s}_2 \odot \mathbf{s}_5.
\end{align}
%
%
As shown, the resulting relationship vector $\mathbf{r}_{A:B}$ approximately expresses the binding between four attribute vectors that are \emph{actively} involved in the relationship.
In fact, $\mathbf{r}_{A:B}$ transparently describes that $\mathbf{t}_{\text{square}}$ should be mapped to $\mathbf{t}_{\text{pentagon}}$, and $\mathbf{s}_2$ should be mapped to $\mathbf{s}_5$.  
This relationship vector provides an explanation as to how the source objects can inductively be mapped to the target objects.
Therefore the relationship can be readily applied beyond the example from which it is learned. 
In order to apply the relationship $\mathbf{r}_{A:B}$ in the target domain $(\alpha: \beta)$, we bind the relationship vector with the object vector $\mathbf{q}_{\alpha}$, resulting in:
\begin{align}
    \hat{\mathbf{o}}_{\beta} =  \mathbf{r}_{A:B}  \odot \mathbf{q}_{\alpha} \approx 
    \\
    \mathbf{r}_{A:B}  \odot \mathbf{o}_{\alpha} =
    \\
    (\mathbf{t}_{\text{square}} \odot \mathbf{t}_{\text{pentagon}} \odot \mathbf{s}_2 \odot \mathbf{s}_5) \odot (\mathbf{l}_2 \odot \mathbf{t}_{\text{square}} \odot \mathbf{s}_2 \odot \mathbf{c}_3) =
    \\
    \mathbf{l}_2 \odot \mathbf{t}_{\text{pentagon}} \odot \mathbf{s}_5 \odot \mathbf{c}_3 = \mathbf{o}_{\beta}.
\end{align}
We observe that the VSA-generated object ($\hat{\mathbf{o}}_{\beta}$) matches the target object ($\mathbf{o}_{\beta}$).
%
In short, the final answer generation consists of binding both panels from the source domain and the first panel from the target domain, i.e., $\hat{\mathbf{o}}_{\beta} = \mathbf{q}_{A} \odot \mathbf{q}_{B} \odot \mathbf{q}_{\alpha}$.

We generated our \texttt{one\_to\_one} problems by using the 2x2 grid constellation of the RAVEN dataset, where we used the rules \texttt{constant} and \texttt{distribute\_two} in the generation process. 
The rules are applied to the position, type, size, and color. 
The \texttt{constant} rule fixes the value of an attribute in a row, whereas in the \texttt{distribute\_two} rule, we sample two valid and distinct values and assign the first value to the object in panels $A$ and $\alpha$ and the second value to panels $B$ and $\beta$. 
The rule of each attribute is selected individually, with the constraint that there is at least one attribute with the \texttt{constant} rule. 
The attribute values of the object in panel $\alpha$ are only allowed to differ if the rule on the attribute is \texttt{constant}. 
We generated 6,000 \texttt{one\_to\_one} problems, where these problems can be further divided into sets of size 2,000 that contain the \texttt{distribute\_two} rule once, twice, and three times.

For evaluation, we trained the \name frontend for 100 epochs with a batchsize of 256 and scaling factor $s_l = 1$ on the panels from the standard RAVEN training set in the 2x2 grid constellation. 
We have used a learning rate of 0.1, which we have decreased by a factor of 10 every 30 epochs. 
We consider the analogy to be successfully solved when the ground-truth VSA representation of the object in panel $\beta$ (i.e., $\mathbf{o}_{\beta}$) has the highest cosine similarity with our generated answer $\hat{\mathbf{o}}_{{\beta}}$.
This has been done by an associative memory cleanup that computes the similarities between the generated vector and all the object vectors in the dictionary $\mathbf{W}$, i.e., $ \mathbf{o}_{\beta} = \mathrm{argmax}_{\mathbf{w}\in \mathbf{W}} \mathrm{ sim}(\hat{\mathbf{o}}_{{\beta}},\mathbf{w})$.
Using the output of our \name frontend, we solved the generated \texttt{one\_to\_one} analogies with an accuracy of 100\%. 

\subsubsection*{b. One to many relationship}
We expand the \texttt{one\_to\_one} analogies to \texttt{one\_to\_many} analogies, where an example is shown in Supplementary Fig.~\ref{fig:one-to-many}. 
The main difference compared to the \texttt{one\_to\_one} analogies is that the number of objects in the panels $B$ and $\beta$ can be more than one, e.g., 2, 3, or 4. 
This means there are relationships between the objects (depicted with different colors in Supplementary Fig.~\ref{fig:one-to-many}).

We describe capturing the relationship set and transferring the relationship set to the target domain based on an example with a fixed number of objects ($k$) in the panels $B$ and $\beta$.
Since there are multiple ($k$) objects in the panel $B$, its ground-truth VSA representation is the addition (i.e., bundling) of the VSA representations of the individual objects ($\mathbf{o}_{B_1}$, ..., $\mathbf{o}_{B_k}$) present in the panel, which is expressed by 
\begin{align}
    \textbf{o}_B = \textbf{o}_{B_1} \oplus ... \oplus \textbf{o}_{B_k}. 
\end{align}
Similarly, the ground-truth VSA representation of the panel $\beta$ is:
\begin{align}
    \textbf{o}_{\beta} = \textbf{o}_{\beta_1} \oplus ... \oplus \textbf{o}_{\beta_k}.
\end{align}
The trained ResNet-18 generates the perceptual representations for the three panels: $\mathbf{q}_A$, $\mathbf{q}_B$, $\mathbf{q}_\alpha$.  
Similar to Supplementary Note~2a, binding $\mathbf{q}_A$ and $\mathbf{q}_B$ captures the relationship set in $\mathbf{r}_{A:B}$.
This is because, in VSA, multiplication (binding) distributes over addition (bundling).
The relationship vector $\mathbf{r}_{A:B}$ is an approximation of the bundle of all \texttt{one\_to\_one} analogies present in the source domain, namely:
\begin{align}
    \mathbf{r}_{A:B} = \textbf{q}_{A} \odot \textbf{q}_b \approx \mathbf{r}_{A:B}^1 \oplus ... \oplus \mathbf{r}_{A:B}^k, \text{where } 
    \mathbf{r}_{A:B}^i = \mathbf{o}_{A} \odot \mathbf{o}_{B_i} \text{ for $i = 1,..., k$.}
\end{align}

Finally, the relationship vector $\mathbf{r}_{A:B}$ is bound by the object representation in panel $\alpha$ to represent the target domain ($\hat{\mathbf{o}}_{\beta}=\mathbf{r}_{A:B} \odot \mathbf{o}_{\alpha}$).
This single binding operation performs computation-in-superposition by applying a set of relationships simultaneously.
We measure the accuracy of our analogy by calculating the cosine similarity between the ground-truth object representation $\mathbf{o}_{\beta}$ and our generated representation $\hat{\mathbf{o}}_{\beta}$. We consider the analogy to be solved when the cosine similarity between the two vectors is at least 0.99.

We generated the \texttt{one\_to\_many} problems similar to the \texttt{one\_to\_one} problems, except that multiple objects are required in panels $B$ and $\beta$. Due to this exception, the position attribute is only allowed to have the \texttt{distribute\_two} rule since the \texttt{constant} rule could not be instantiated. We generated 2000 \texttt{one\_to\_many} problems.

In the experiments, we used the same \name frontend used in Supplementary Note~2a. 
We achieved 100\% accuracy in solving the \texttt{one\_to\_many} visual analogy tasks.

\begin{figure}[!ht]
\centering
    \subfloat[\texttt{one\_to\_one} analogy]{\includegraphics[width=.3\textwidth]{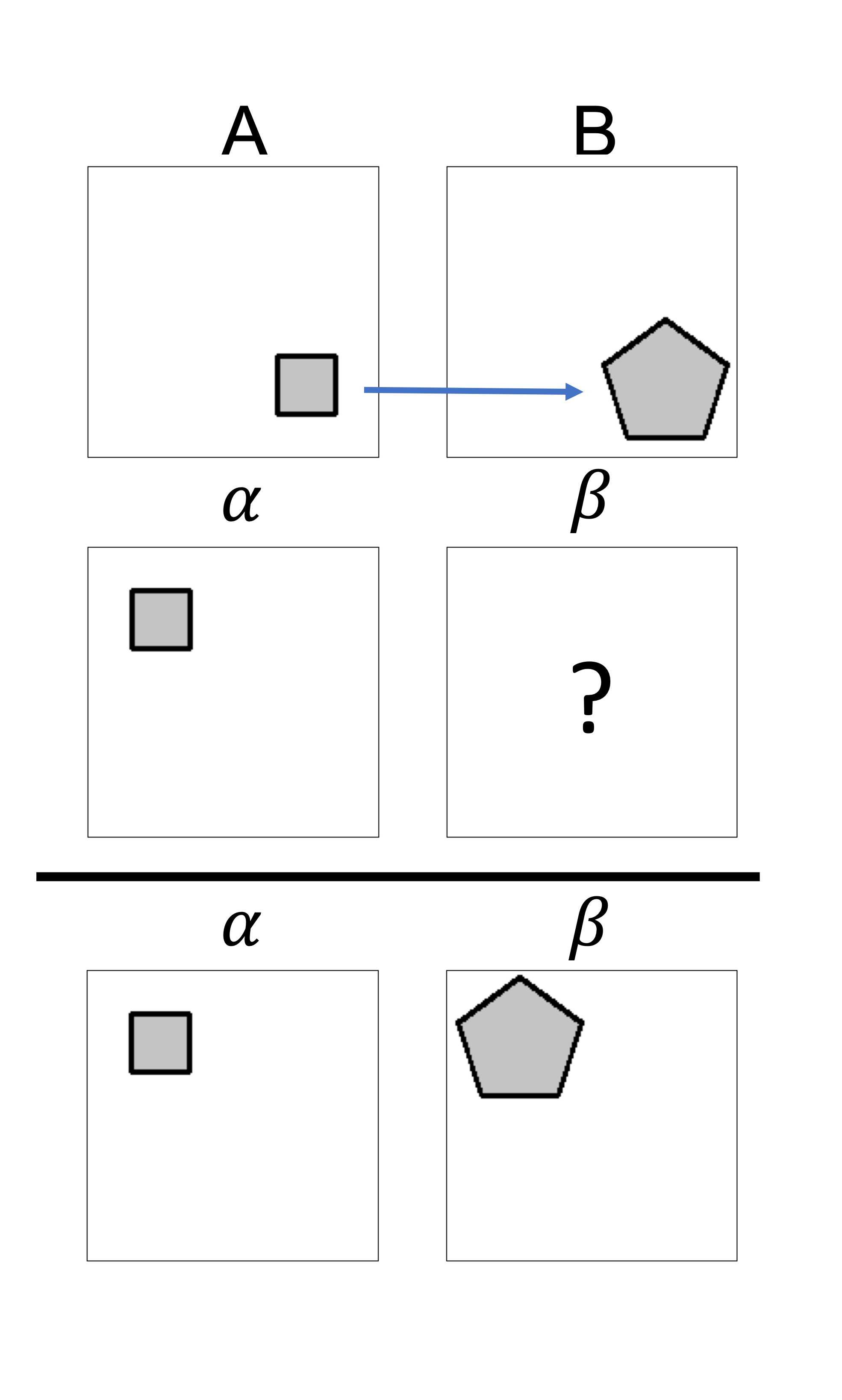}\label{fig:one-to-one}}
    \subfloat[\texttt{one\_to\_many} analogy]{\includegraphics[width=.3\textwidth]{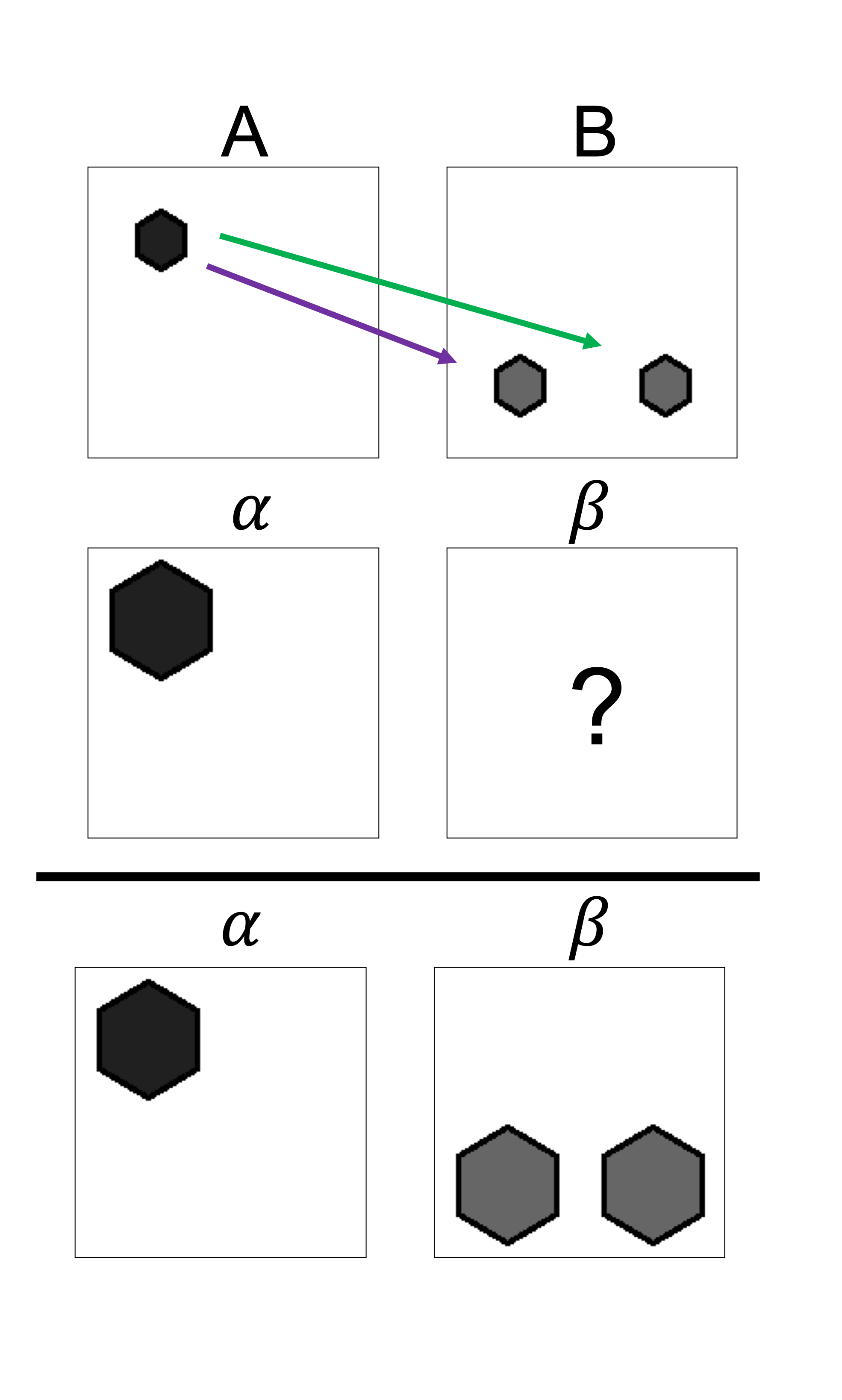}\label{fig:one-to-many}}
\caption{\textbf{Example analogy in (a) \texttt{one\_to\_one} scenario and (b) \texttt{one\_to\_many} scenario.} In \textbf{(a)} we illustrate a \texttt{one\_to\_one} visual analogy problem. In the source domain ($A:B$) there is an underlining relationship of changing type and size of the object which is shown by the blue arrow. The same relationship should be applied to novel circumstances in target domain ($\alpha:\beta$), where for example the position of the objects is different.
In \textbf{(b)} we illustrate expanded \texttt{one\_to\_many} visual analogy problem. In this problem, there is a set of relationships between the objects in the source domain ($A:B$). In the shown example, the relationship set consists of two \texttt{one\_to\_one} relationships indicated by the green an violet arrows. Both relationships change the color of the object in the same way, however they alter the position attribute to a different value.
} 
\end{figure}

\clearpage
\subsection*{Supplementary Note 3: Details on the \name backend}
In this Supplementary Note, we provide a detailed description of the \name backend, including the VSA representation of PMFs, the rule probability computation and execution, and the selection of the rule and the final answer. 
\subsubsection*{VSA representation of PMFs}
For all attributes, the PMF is represented through the weighted superposition with the values in the PMF used as weights and the corresponding codewords $\mathbf{b}_k$ as basis vectors taken from the codebook $B:= \lbrace \mathbf{b}_k \rbrace_{k=1}^{n}$:
{\cc{
\begin{align}\label{eq:bsbc_superpos}
    \mathbf{a}^{(i,j)}:=g(\mathbf{p}^{(i,j)}) =  \sum_{k=1}^{n} \mathbf{p}^{(i,j)}[k] \cdot \mathbf{b}_k.
\end{align}
}}
The codebook (discrete or continuous) is selected based on the underlying attribute and rule. 
The number of codewords $n$ is given by the dimensionality of the PMF, which depends on the attribute.
For example, the shape PMF is 5-dimensional due to the five different shapes in RAVEN; hence, the transformation requires $n=5$ codewords. 
After transforming the PMF to a VSA representation, we can manipulate the PMFs in the VSA representation using the algebra provided by the vector space. 
This allows estimating the probability $\mathbf{u}[\texttt{rule}]$ for every \texttt{rule}. 
Then, the most probable rule is selected and executed in the vector space, yielding $\mathbf{\hat{a}}^{(3,3)}$.
{\cc{
Finally, the PMF is estimated using the similarity computation with a consecutive normalization: 
\begin{align}
    \mathbf{\hat{p}}^{(3,3)}:= \mathrm{norm}\left(  \left[\mathrm{sim}(\mathbf{\hat{a}}^{(3,3)},\mathbf{b}_1), \mathrm{sim}(\mathbf{\hat{a}}^{(3,3)},\mathbf{b}_2),..., \mathrm{sim}(\mathbf{\hat{a}}^{(3,3)},\mathbf{b}_n),  \right] \right).
\end{align}
}}

\begin{figure}[!ht]
\centering
    \subfloat[Discrete]{\includegraphics[width=.45\textwidth]{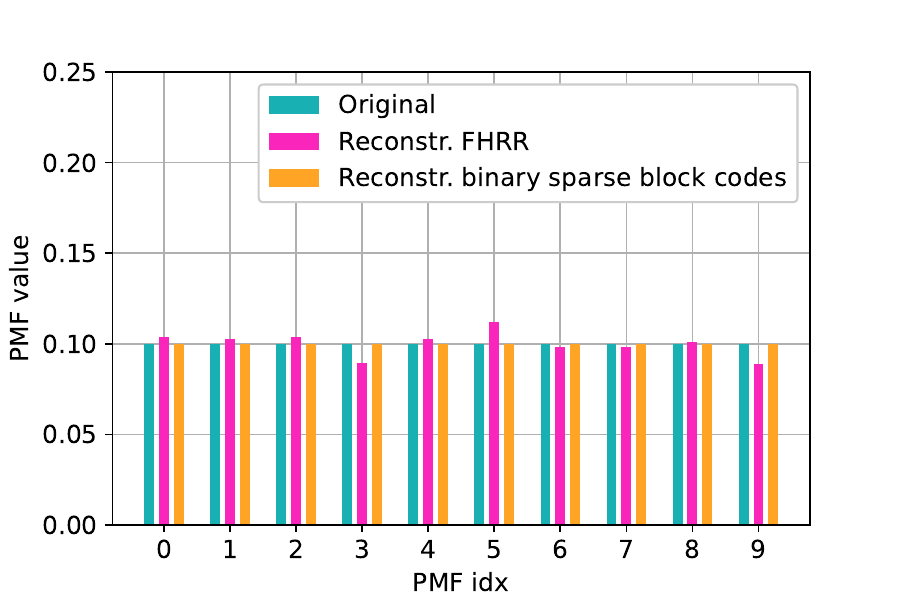}\label{fig:discrete}}
    \subfloat[Continuous]{\includegraphics[width=.45\textwidth]{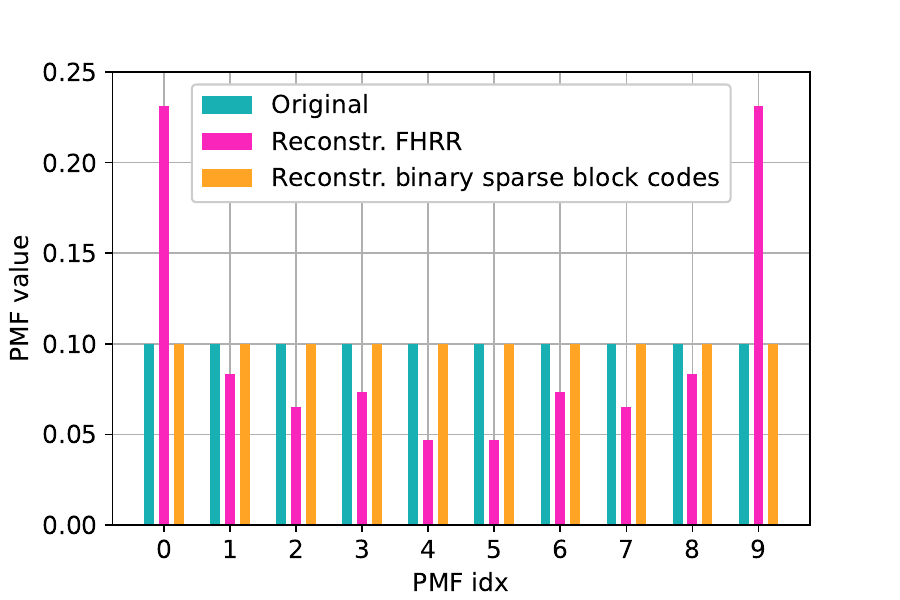}\label{fig:continuous}}
\caption{\cc{Reconstruction of uniformly distributed PMF using either FHRR or binary sparse block codes for both discrete and continuous concepts. The dimension of both VSA representations is set to $d$=1024.}} \label{fig:reconstruction}
\end{figure}

{\cc{
An alternative VSA representation to the binary sparse block codes is Fourier holographic reduced representation (FHRR)~\cite{PlateHolographic2003}. 
In the following, we describe a similar procedure of transforming a PMF to an FHRR-based VSA representation and compare it with the binary sparse block codes.
The basis vectors in FHRR are $d$-dimensional, complex-valued, unary vectors. 
Each element is a complex phasor with unit norm and an angle randomly drawn from a uniform distribution $U(-\pi,\pi)$. 
The dense bipolar representations are a particular case of the FHRR model where angles are restricted to $\lbrace 0, \pi \rbrace$. 
The binding in FHRR is defined as the element-wise modulo-$2\pi$ sum; similarly, the unbinding is the element-wise modulo-$2\pi$ difference. 
The bundling of two or more vectors is computed via the element-wise addition with a consecutive normalization step, which sets the magnitude of each phasor to unit magnitude. 
The similarity of two vectors is the sum of the cosines of the differences between the corresponding angles.
Binding, unbinding, and similarity computation can be done using the polar coordinates, while bundling requires the Cartesian coordinates.
For a discrete attribute, we use a codebook with $n$ unrelated basis vectors $\mathbf{b}_i \in \mathbb{C} ^ {d}$. 
For representing the PMF of a continuous attribute, we use a codebook with basis vectors generated by the fractional power encoding~\cite{PlateHolographic2003}, where the basis vector corresponding to an attribute value $v$ is defined by exponentiation of a randomly chosen basis vector $\mathbf{e}$ using the value as the exponent, i.e., $\mathbf{b}_v = \mathbf{e}^v$.
Each PMF is represented through the normalized weighted superposition with the values in the PMF used as weights and the corresponding codewords as basis vectors:
\begin{align}\label{eq:fhrr_superpos}
    \mathbf{a}_{\text{FHRR}}^{(i,j)}:=g_{\text{FHRR}}(\mathbf{p}^{(i,j)}) = \mathrm{cnorm}\left( \sum_{k=1}^{n} \mathbf{p}^{(i,j)}[k] \cdot \mathbf{b}_k \right),  
\end{align}
where $\mathrm{cnorm}(\cdot)$ normalizes the magnitude of every phasor of a $d$-dimensional complex-valued vector. 

However, we observed nonidealities when mapping PMFs to the FHRR-based VSA representations. 
In a synthetic experiment, we mapped a uniform distribution to the VSA representation using either the binary sparse block codes (by equation~\eqref{eq:bsbc_superpos}) or FHRR (by equation~\eqref{eq:fhrr_superpos}), and projected them back to the PMF representation again using the associative memory search on the corresponding codebook. 
Supplementary Fig.~\ref{fig:reconstruction} shows the original and the reconstructed PMFs for discrete and continuous concepts.
We observe that both FHRR and binary sparse block codes can represent the discrete PMFs; however, FHRR faces issues in representing continuous PMFs. 
This nonideality might stem from the complex normalization step in combination with the constructive interference of the superimposed complex phasors.
Thus, in this work, we use sparse binary block codes in our \name backend. 
}}

\subsubsection*{Rule probability computation and rule execution}
In the following, we describe the probability computation and the execution for every rule. 

\paragraph{Arithmetic plus and minus.}
The arithmetic rule computes the value of the last panel by adding (arithmetic plus) or subtracting (arithmetic minus) the attribute values of the first two panels. 
Since this rule relates to a continuous concept, we use a dictionary constructed by fractional power encoding. 
For determining the rule probability for arithmetic plus, we represent the addition of the first two panels using the binding operation
\begin{align}\label{eq:arithmetic_plus}
     \mathbf{r}^{+}_{i} = \mathbf{a}^{(i,1)} \odot \mathbf{a}^{(i,2)}, \quad i\in \lbrace 1,2 \rbrace. 
\end{align}
If the arithmetic plus rule applies, we expect $\mathrm{sim}( \mathbf{r}^+_{i},\mathbf{a}^{(i,3)} )\gg 0$. 
Let us have a closer look at the similarity expression for the first row: 
\begin{align}
    \mathrm{sim}( \mathbf{r}^+_{i},\mathbf{a}^{(1,3)} ) &=  \mathrm{sim}(\mathbf{a}^{(1,1)}\odot  \mathbf{a}^{(1,2)} ,\mathbf{a}^{(1,3)} ) \label{eq:deriv0}\\
    &= \mathrm{sim}\left( \left( \sum_{k_1=1}^{n} \mathbf{p}^{(1,1)}[k_1] \cdot \mathbf{b}_{k_1} \right) \odot \left(\sum_{k_2=1}^{n} \mathbf{p}^{(1,2)}[k_2] \cdot \mathbf{b}_{k_2} \right), \left( \sum_{k_3=1}^{n} \mathbf{p}^{(1,3)}[k_3] \cdot \mathbf{b}_{k_3} \right) \right)\label{eq:deriv2}\\
    &= \sum_{\substack{k_1, k_2, k_3\\  s.t. k_1+k_2=k_3}} \mathbf{p}^{(1,1)}[k_1] \cdot \mathbf{p}^{(1,2)}[k_2] \cdot \mathbf{p}^{(1,3)}[k_3] + \sum_{\substack{k_1, k_2, k_3\\  s.t. k_1+k_2\neq k_3}} \mathrm{sim}(\mathbf{b}_{k_1}\odot \mathbf{b}_{k_2}, \mathbf{b}_{k_3})\mathbf{p}^{(1,1)}[k_1] \cdot \mathbf{p}^{(1,2)}[k_2] \cdot \mathbf{p}^{(1,3)}[k_3] \label{eq:deriv3}
    \\
    &= \sum_{\substack{k_1, k_2, k_3\\  s.t. k_1+k_2=k_3}} \mathbf{p}^{(1,1)}[k_1] \cdot \mathbf{p}^{(1,2)}[k_2] \cdot \mathbf{p}^{(1,3)}[k_3] + \sum_{\substack{k_1, k_2, k_3\\  s.t. k_1+k_2\neq k_3}} n_{k_1,k_2,k_3}\cdot\mathbf{p}^{(1,1)}[k_1] \cdot \mathbf{p}^{(1,2)}[k_2] \cdot \mathbf{p}^{(1,3)}[k_3] \label{eq:deriv4}\\
    &\approx \sum_{\substack{k_1, k_2, k_3\\  s.t. k_1+k_2=k_3}} \mathbf{p}^{(1,1)}[k_1] \cdot \mathbf{p}^{(1,2)}[k_2] \cdot \mathbf{p}^{(1,3)}[k_3] \label{eq:deriv5}.
\end{align}
%
%
Equation~\eqref{eq:deriv3} uses the linearity of the similarity and divides the sum into contributions that satisfy the arithmetic plus constraint (LHS), i.e.,  $\mathrm{sim}(\mathbf{b}_{k_1} \odot \mathbf{b}_{k_2}, \mathbf{b}_{k_3})=1$, and contributions which do not satisfy the constraint (RHS). 
For the latter, we replace the similarity $\mathrm{sim}(\mathbf{b}_{k_1} \odot \mathbf{b}_{k_2},\mathbf{b}_{k_3})$ with $n_{k_1,k_2,k_3}$, which can be modeled as vanishing noise as the dimension $d$ increases.
The non-satisfying terms converge to zero with sufficiently large dimension $d$ and we can approximate~\eqref{eq:deriv4} with~\eqref{eq:deriv5}. 
Equation~\eqref{eq:deriv5} sums up the products of all valid rule implementations. 
Indeed, this computation appears in traditional probabilistic reasoning engines such as the PrAE~\cite{PrAE_CVPR21}.
{\ccc{
Supplementary Table~\ref{tab:rules} shows the relation between our \name backend and the PrAE backend~\cite{PrAE_CVPR21} for computing the probabilities for different rules.  
Our \name backend derives the rule probability based on the similarity between vectors of fixed dimension, while the PrAE computes the rule probability by marginalizing all possible rule implementations. 
}}

For interpreting the similarity in equation \eqref{eq:deriv0} as a probability, we limit the range of the similarity using a threshold function, which sets all similarity values below 0.05 to 0. 
This suppresses noise stemming from invalid contributions. 

Some of the rules need to satisfy additional constraints to be valid. 
For the arithmetic plus, the sum of the attributes of the first two panels ($k_1+k_2$) has to be smaller than $n$.
By computing $\mathrm{sim}( \mathbf{r}^+_{i},\mathbf{a}^{(i,3)} )$ for $i\in\lbrace 1,2\rbrace$, this constraint is embedded for the first two rows. 
For validating the arithmetic rule in the last row, we compute the constraint 
\begin{align}
    h_a(\mathbf{a}^{(3,1)},\mathbf{a}^{(3,2)}) := \mathrm{min}\left(\sum_{k=1}^n \mathrm{sim}\left( \mathbf{a}^{(3,1)}\odot  \mathbf{a}^{(3,2)}, \mathbf{b}_k \right), 1\right). 
\end{align}
The constraint accumulates all projections of the binding of $\mathbf{a}^{(3,1)}$ and $ \mathbf{a}^{(3,2)}$ (i.e., the addition) to the space spanned by $B$. 
The $\mathrm{min}(\cdot,1)$ guarantees the constraint to be $0\leq h_a \leq 1$ such that it can be interpreted as a probability. 
If the majority of the binding falls outside of the space, i.e., the addition is larger than $n$, the constraint is not satisfied, and its value will be close to zero. 
Finally, the rule probability is determined as 
\begin{align}
    \mathbf{u}[\texttt{arithmetic plus}] = \mathrm{sim}( \mathbf{r}^+_{1},\mathbf{a}^{(1,3)} ) \cdot \mathrm{sim}( \mathbf{r}^+_{2},\mathbf{a}^{(2,3)} ) \cdot h_a(\mathbf{a}^{(3,1)},\mathbf{a}^{(3,2)}).
\end{align}
If the rule is selected, it is executed by computing 
\begin{align}
    \mathbf{\hat{a}}^{(3,3)} = \mathbf{a}^{(3,1)} \odot \mathbf{a}^{(3,2)}.
\end{align}

The rule arithmetic minus is implemented analogously, where the row representation is computed using the unbinding operation:
\begin{align}\label{eq:aminusdet}
     \mathbf{r}^-_{i} = \mathbf{a}^{(i,1)}  \circledast \mathbf{a}^{(i,2)}, \quad i\in \lbrace 1,2 \rbrace. 
\end{align}

\paragraph{Progression.}
{\cc{
The progression rule describes a positive or negative increment by one or two values along the panels; hence, it is a continuous concept, too. 
We detect and execute the progression rules with different increments and decrements individually.
The RAVEN dataset applies the rules row-wise. 
We compute the rule probability for positive increments by computing first the unbinding between adjacent panels \begin{align}
    \mathbf{d}^{+(i,j)} = \mathbf{a}^{(i,j+1)}\circledast \mathbf{a}^{(i,j)} \quad (i, j) \in \lbrace (1,1), (1,2),(2,1),(2,2), (3,1) \rbrace 
\end{align}
as well as the unbinding between the left-most and right-most panel in the first two rows: 
\begin{align}
    \mathbf{d}^{++(i,0)} = \mathbf{a}^{(i,2)}\circledast \mathbf{a}^{(i,0)} \quad i \in \lbrace 1,2 \rbrace. 
\end{align}
If the progression rule by an increment of $n\in \lbrace 1,2 \rbrace$ is active, we expect the unbound vectors $\mathbf{d}^{+(i,j)}$ and $\mathbf{d}^{++(i,0)}$ to be similar to the basis vectors that represent the values $n$ and $2n$ ($\mathbf{b}_n$ and $\mathbf{b}^2_n$):

\begin{align}\label{eq:progression}
    \mathbf{u}[\texttt{progression-plus-}n] = \left(\prod_{i,j} \mathrm{sim}(\mathbf{d}^{+(i,j)}, \mathbf{b}_n) \right) \cdot \left(\prod_{i\in \lbrace 1,2\rbrace} \mathrm{sim}(\mathbf{d}^{++(i,0)}, \mathbf{b}^2_n) \right) \cdot h_p(\mathbf{d}^{(1,1)}). 
\end{align}
The last term prevents us from confusing the progression rule with the constant rule, which can be interpreted as a progression of zero, and is implemented as
\begin{align}
    h_p(\mathbf{d}^{+(1,1)}):=(1-\mathrm{sim}(\mathbf{d}^{+(1,1)}, \mathbf{0})).
\end{align}
It computes the similarity between a difference vector ($\mathbf{d}^{(1,1)}$) and the all-zero vector ($\mathbf{0}$), which is represented with a vector where each block has its non-zero element at index 0. 
Finally, the progression rule is executed by 
\begin{align}\label{eq:progression-exe}
    \mathbf{\hat{a}}^{(3,3)} = \mathbf{a}^{(3,2)} \odot \mathbf{b}_n. 
\end{align}
The implementation of the progression with decrement is analogous, where we compute the binding in reverse order, e.g., 
\begin{align}
    \mathbf{d}^{-(i,j)} = \mathbf{a}^{(i,j)}\circledast \mathbf{a}^{(i,j+1)} \quad (i, j) \in \lbrace (1,1), (1,2),(2,1),(2,2), (3,1) \rbrace. 
\end{align}
}}

\paragraph{Distribute three.}
The distribute three rule relates to a discrete concept, hence we use fully random codewords. 
First, we compute the row-wise binding of the PMF-vector representation of the first two rows
\begin{align}
     \mathbf{r}_{i} = \mathbf{a}^{(i,1)} \odot \mathbf{a}^{(i,2)} \odot \mathbf{a}^{(i,3)}, \quad i\in \lbrace 1,2 \rbrace. 
\end{align}
Similarly, we can compute the column representations by 
\begin{align}
     \mathbf{c}_{j} = \mathbf{a}^{(1,j)} \odot \mathbf{a}^{(2,j)} \odot \mathbf{a}^{(3,j)}, \quad j\in \lbrace 1,2 \rbrace. 
\end{align}
If the distribute three rule applies, we expect both $\mathrm{sim}(\mathbf{r}_1,\mathbf{r}_2)\gg 0$ and $\mathrm{sim}(\mathbf{c}_1,\mathbf{c}_2)\gg 0$. 

{\cc{
Hence, the rule probability is computed by 
\begin{align}
    \mathbf{u}[\texttt{distribute three}] = \mathrm{sim}(\mathbf{c}_1,\mathbf{c}_2) \cdot \mathrm{sim}(\mathbf{r}_1,\mathbf{r}_2)\cdot h_d(\mathbf{a}^{(1,1)},\mathbf{a}^{(1,2)}), ..., \mathbf{a}^{(2,3)}), 
\end{align}
where 
\begin{align}
    h_d(\mathbf{a}^{(1,1)},\mathbf{a}^{(1,2)}), ..., \mathbf{a}^{(2,3)}):= (1-\mathrm{sim}(\mathbf{a}^{(1,1)},\mathbf{a}^{(1,2)})\cdot (1-\mathrm{sim}(\mathbf{a}^{(1,2)},\mathbf{a}^{(1,3)})\cdot ... \cdot (1-\mathrm{sim}(\mathbf{a}^{(3,1)},\mathbf{a}^{(3,2)})
\end{align}
validates the constraint that panels are not equal within a row.
}}
If the rule is selected, it is executed by
\begin{align}
    \mathbf{\hat{a}}^{(3,3)} = \mathbf{r}_1 \circledast \left(\mathbf{a}^{(3,1)} \odot \mathbf{a}^{(3,2)} \right).
\end{align}

\paragraph{Constant.}
The computation of the constant rule probability involves the row-wise similarities: 
\begin{align}
    \mathbf{u}[\texttt{constant}]=\mathrm{sim}(\mathbf{a}^{(1,1)},\mathbf{a}^{(1,2)})\cdot \mathrm{sim}(\mathbf{a}^{(1,2)},\mathbf{a}^{(1,3)})\cdot \mathrm{sim}(\mathbf{a}^{(2,1)},\mathbf{a}^{(2,2)}) \cdot \mathrm{sim}(\mathbf{a}^{(2,2)},\mathbf{a}^{(2,3)})\cdot \mathrm{sim}(\mathbf{a}^{(3,1)},\mathbf{a}^{(3,2)}).
\end{align}

The execution of the constant rule requires no transformation; thus, there is no need to map the PMF to the vector space and back. 
Therefore, the PMF of the missing panel can be directly estimated by using one of the PMFs in the bottom row, e.g., 
\begin{align}
     \mathbf{\hat{p}}^{(3,3)} = \mathbf{p}^{(3,1)}. 
\end{align}

\begin{table}
\caption{\ccc{Relation of rule probability computation between PrAE~\cite{PrAE_CVPR21} and our \name backend.}}\label{tab:rules}
\everymath{\displaystyle}
\resizebox{\textwidth}{!}{%
{\ccc{
\begin{tabular}{l|l|l}
\toprule
Rule & PrAE & NVSA\\
\midrule
\texttt{Constant} & $ u= \sum\limits_{r=1}^2  \sum\limits_{v=1}^n   \prod\limits_{c=1}^3 \mathbf{p}^{(r,c)}[v] + \sum\limits_{v=1}^n   \prod\limits_{c=1}^2 \mathbf{p}^{(3,c)}[v] $& $u=\mathrm{sim}(\mathbf{a}^{(1,1)},\mathbf{a}^{(1,2)})\cdot \mathrm{sim}(\mathbf{a}^{(1,2)},\mathbf{a}^{(1,3)})\cdot ... \cdot \mathrm{sim}(\mathbf{a}^{(3,1)},\mathbf{a}^{(3,2)})$ \\
\midrule
\begin{tabular}[c]{@{}l@{}}\texttt{Progression-} \\ \texttt{plus n}\end{tabular} & $ u= \sum\limits_{r=1}^2  \sum\limits_{\substack{v_1, v_2, v_3\\  s.t.\, v_1+n=v_2 \\
v_2+n = v_3}}   \prod\limits_{c=1}^3 \mathbf{p}^{(r,c)}[v_c] + \sum\limits_{\substack{v_1, v_2\\  s.t.\, v_1+n=v_2}}   \prod\limits_{c=1}^2 \mathbf{p}^{(3,c)}[v_c] $& $u=\left(\prod_{i,j} \mathrm{sim}(\mathbf{d}^{+(i,j)}, \mathbf{b}_n) \right) \cdot \left(\prod_{i\in \lbrace 1,2\rbrace} \mathrm{sim}(\mathbf{d}^{++(i,0)}, \mathbf{b}^2_n) \right) \cdot h_p(\mathbf{d}^{(1,1)})$ \\
\midrule
\begin{tabular}[c]{@{}l@{}}\texttt{Arithmetic-} \\ \texttt{plus}\end{tabular} & $ u= \sum\limits_{r=1}^2  \sum\limits_{\substack{v_1, v_2, v_3\\  s.t.\, v_1+v_2=v_3}}   \prod\limits_{c=1}^3 \mathbf{p}^{(r,c)}[v_c]$ & $u=\mathrm{sim}( \mathbf{r}^+_{1},\mathbf{a}^{(1,3)} ) \cdot \mathrm{sim}( \mathbf{r}^+_{2},\mathbf{a}^{(2,3)} ) \cdot h_a(\mathbf{a}^{(3,1)},\mathbf{a}^{(3,2)})$ \\
\midrule
\texttt{Dist.three} & $ u= \sum\limits_{\substack{v^{(1,1)},...,v^{(3,2)} \in I_{d3}}} \prod\limits_{r=1}^2 \prod\limits_{c=1}^3 \mathbf{p}^{(r,c)}[v^{(r,c)}]\prod\limits_{c=1}^2 \mathbf{p}^{(3,c)}[v^{(3,c)}]$ & $u=\mathrm{sim}(\mathbf{c}_1,\mathbf{c}_2) \cdot \mathrm{sim}(\mathbf{r}_1,\mathbf{r}_2)\cdot h_d(\mathbf{a}^{(1,1)},\mathbf{a}^{(1,2)}), ..., \mathbf{a}^{(2,3)})$ \\
\bottomrule
\end{tabular}
}}
}
\end{table}

\subsubsection*{Selection of the rule and the final answer}
For each attribute, we compute the rule probability of all rules using the \name backend. 
In the RAVEN dataset, the arithmetic and progression rules on the position attribute are implemented in the binary system; thus, we compute the rule probability computation in the original PMF space for those rules.
For every attribute, we select the rule with the highest rule probability and apply it to get $\hat{P}^{(3,3)}:= (\mathbf{\hat{p}}_{\text{pos}},\mathbf{\hat{p}}_{\text{num}}, \mathbf{\hat{p}}_{\text{type}},\mathbf{\hat{p}}_{\text{size}}, \mathbf{\hat{p}}_{\text{color}}$).
Finally, we compute for each candidate answer panel $j$, the Jensen–Shannon divergence (JSD) between each of the five probability distributions in $P^{(k)}$ and $\hat{P}^{(3,3)}$, and sum the five JSD values to obtain a score for the answer panel $j$. 
The predicted answer panel $j^{\star}$ is the one with the lowest total divergence.

\clearpage
\subsection*{Supplementary Note 4: Out-Of-Distribution generalization to unseen attribute-rule pairs}
\cc{
In this Supplementary Note, we evaluate the generalizability of our \name, including both frontend and backend, for to unseen attribute-rule pairs. 
More specifically, we evaluate whether our model is able to solve an unseen target attribute-rule pair (e.g., the constant rule on the type attribute) when it has been trained on the examples containing all of the attribute-rule pairs except the specific target one (e.g., the constant rule on size and color, the progression rule on all attributes, and the distribute rule on all attributes). 
Hence, this setting tests the out-of-distribution generalization for the attribute-rule pairs.
To do so, we generate a new training and validation set containing all examples except those with the target attribute-rule pair and a test set containing examples exclusively with the target attribute-rule pair.  
The datasets are generated by filtering the existing splits in RAVEN and I-RAVEN. 
As a result, the sets contain fewer samples depending on the target attribute-rule pair; specifically, the training sets contain 2622--3437 samples, the validation sets 841--1160 samples, and the test sets 803--1117.

Supplementary Table~\ref{tab:results-ruleattribute-generalization} shows the experimental results on I-RAVEN in the L-R constellation. 
The results of LEN~\cite{zheng2019abstract} and CoPINet~\cite{CoPINET_19} are based on experiments conducted by Wu \textit{et al.}~\cite{wu2020scl}. 
Our \name was trained end-to-end and outperformed both baselines by a large margin in all target attribute-rule pairs. 
Minor accuracy degradations are observed in the continuous rules (i.e., progression and arithmetic). 
This might point out the importance of the continuous rules being present for the \name to learn all attribute values properly. 
}

\begin{table}[h]
\caption{\cc{Out-Of-Distribution generalization on the unseen rule-attribute pairs of the I-RAVEN dataset in the L-R constellation. We report accuracy (\%) on test set that contains exclusively examples with the target attribute-value pairs on which it has not been trained on.} 
}\label{tab:results-ruleattribute-generalization}
{\cc{
\begin{tabular}{lrrrrrrrrrrr}
\toprule
        & \multicolumn{3}{c}{Type} & \multicolumn{4}{c}{Size}  & \multicolumn{4}{c}{Color} \\
        \cmidrule(r){2-4} \cmidrule(r){5-8} \cmidrule(r){9-12}
        & Constant      & Progress.      & Dist.3      & Constant      & Progress.      & Dist.3 & Arithmetic    & Constant      & Progress.      & Dist.3 & Arithmetic  \\
        \cmidrule(r){1-1} \cmidrule(r){2-4} \cmidrule(r){5-8} \cmidrule(r){9-12}
LEN\cite{zheng2019abstract}    & 28.0   & 24.0    & 29.4   & 24.4  & 27.9  & 27.6 & -    & 25.3 & 25.3 & 22.0 & -    \\
CoPINet\cite{CoPINET_19} & 25.1    & 36.2   & 32.9   & 37.2 & 36.2 & 36.4 &  -    & 38.8 & 35.8  & 29.2  & -    \\
NVSA    & 100     & 81.8   & 100     & 100  & 100   & 100  & 77.8  & 99.9  & 81.7  & 100   & 80.9  \\
\bottomrule
\end{tabular}
}}
\end{table}

\newpage
\subsection*{Supplementary Note 5: Experiments on the PGM dataset}
{\ccc{
This section describes the application of our \name to the Procedurally Generated Matrices (PGM) dataset~\cite{barrett2018measuring}.
}}

\subsubsection*{PGM Dataset}
{\ccc{
The PGM dataset provides RPM tests with two constellations, line and shape, which can simultaneously be present in a panel.
The objects in the shape constellation are arranged in a 3x3 grid, each taking one out of ten gray shadings, ten sizes, and seven geometrical types (see Supplementary Table~\ref{tab:pgm-attribute}).
Moreover, the line constellation contains six different line types, each taking one out of ten gray shadings. 
Both constellations can have no objects present in a panel. 

Each PGM example has 1 to 4 active rules, either applied row-wise or column-wise. 
This contrasts the RAVEN dataset, where only a row-wise rule governs every attribute.
The rules can be described as follows: 
\begin{itemize}
    \item \texttt{Progression}: The attribute value monotonically increases by a value of one in a row/column. 
    \item \texttt{XOR}, \texttt{OR}, and \texttt{AND}: The set of attribute values in the third panel in a row/column corresponds to the logical \texttt{XOR}, \texttt{OR}, or \texttt{AND} operation of the first two panels. Let us consider an example with the attribute type in the shape constellation. The first panel contains objects with squares and triangles and the second only triangles. Consequently, the third panel would either contain only squares (\texttt{XOR}), both squares and triangles (\texttt{OR}), or only triangles (\texttt{AND}).
    \item \texttt{Consistent union}: The same set of attribute values appear in the three panels of every row/column (with permutations of the values in different rows/columns). This is a relaxed version of the \texttt{distribute three} rule in the RAVEN dataset since it does not require all the permutations to be distinct. 
\end{itemize}

Supplementary Table~\ref{tab:pgm-attribute} summarizes the attribute rules of the two constellations.
The PGM dataset contains 1,200,000 examples for training, 20,000 for validation, and 200,000 for testing. 
Moreover, the active rules are provided as meta-labels. 
Note that the meta-labels do not contain the orientation of the rule (i.e., row-wise or column-wise).  
}}

\subsubsection*{\name frontend}
We start by defining the codebooks for the two constellations. 
The line constellation has two codebooks, $T_L:=\{\mathbf{t}_i\}_{i=1}^6$ for type and $C_L:=\{\mathbf{c}_i\}_{i=1}^{10}$ for color. 
Similarly, the codebooks for the shape constellation are $T_S:=\{\mathbf{t}_i\}_{i=1}^7$, $S_S:=\{\mathbf{s}_i\}_{i=1}^{10}$,  $C_S:=\{\mathbf{c}^S_i\}_{i=1}^{10}$, and $L_S:=\{\mathbf{l}_i\}_{i=1}^{9}$ representing the type, size, color, and position of a single object. 
We build the dictionaries $\mathbf{W}_L \in \Bipolar^{m_L \times d}$ and $\mathbf{W}_S \in \Bipolar^{m_S \times d}$ by binding the vectors from the codebooks of all possible combinations for line and shape. 
This yields $m_S$=6300 combinations for the shape and $m_L$=60 combinations for the line. 

In our earlier \name frontend using one ResNet-18, we identified a limitation in the end-to-end training with multiple attributes, where in most cases, only one attribute was learned, and the others remained at random chance.
The limitation might stem from the larger number of attribute combinations in PGM: the shapes in PGM have $>2\times$ more attribute combinations than the largest 3x3 grid constellation in RAVEN (6300 vs. 2700). 
To simplify the end-to-end training, we train four ResNet-18 models, each focusing on two attributes: one for the line constellation (type-color attributes) and three for the shape constellation (type-position, size-position, color-position).
We intentionally included the position attribute for every attribute since it influences the scene probability computation of the other attributes (e.g., see equation~(17) in Methods). 
Moreover, we use a larger dimension $d=1024$ for better performance.

\subsubsection*{Probabilistic scene representation}

For every panel, we compute a PMF for each object (e.g., $\mathbf{v}^{(k)}_{\text{exist}}$, $\mathbf{v}^{(k)}_{\text{type}}$, $\mathbf{v}^{(k)}_{\text{size}}$, $\mathbf{v}^{(k)}_{\text{color}}$ for object $k$ in the shape constellation) using the marginalization with consecutive softmax approach (see equation~(12)--(14) in Methods). 
We then derive the PMFs representing the attributes of the panel. 
Here, most attribute-rule pairs use the same computation of the panel PMF as in RAVEN.
The exceptions are the logical rules (\texttt{XOR}, \texttt{OR}, and \texttt{AND}) on color, size, and type, where we need to describe every attribute value combination separately. 
More specifically, we describe the occupancy with the set of occupied values $I_j$; e.g., $I_3=\{1,2\}$ represents a scene with at least one object with attribute value 1 and at least one with attribute value 2. 
The probability that a panel contains the attributes $a$ with values in $I_j$ is determined by
\begin{align}\label{eq:ext-PMF}
    \mathbf{p}'_a[j] = \prod_{k\in I_j} \mathrm{min}\left(\sum_{l=1}^n v^{(l)}_{\text{exist}}[0] v^{(l)}_a[k] , 1 \right), 
\end{align}
where $n$ is the number of positions in the scene ($n$=6 for line and $n$=9 for shape), $v_{\text{exist}}^{(l)}$ the probability that an object exists at position $l$, and $v_a^{(l)}[k]$ the probability that the object at position $l$ has attribute $a$ with value $k$.  
We limit the set $I_j$ to contain at most four different values to keep the compute and memory demands low.
Overall, we get a scene representation for the line, $P_L:=\{ \mathbf{p}_{\text{pos}},  
\mathbf{p}_{\text{color}}, \mathbf{p}'_{\text{color}}\}$, and for the shape constellation, $P_S:=\{ \mathbf{p}_{\text{pos}}, \mathbf{p}_{\text{num}}, \mathbf{p}_{\text{type}}, \mathbf{p}'_{\text{type}}, \mathbf{p}_{\text{size}}, \mathbf{p}'_{\text{size}},  \mathbf{p}_{\text{color}}, \mathbf{p}'_{\text{color}}\}$.  
Note that the attributes color, type, and size now have two scene representations: the standard PMF $\mathbf{p}_a$ (see equation~(17) in Methods) and the novel extended $\mathbf{p}'_a$ (see equation~\eqref{eq:ext-PMF}). 
As opposed to the RAVEN dataset, an inconsistency state is not required for the PGM dataset. 

\begin{table}[]
\centering
\caption{Summary of attributes and rules in the PGM dataset.}
\label{tab:pgm-attribute}
\begin{tabular}{llrl}
\toprule
Constellation & Attribute name & \begin{tabular}[c]{@{}c@{}}Number of \\ attribute values\end{tabular} & \multicolumn{1}{c}{Rules}                   \\
\cmidrule(r){1-1}\cmidrule(r){2-2}\cmidrule(r){3-3}\cmidrule(r){4-4}
Shapes        & Color          & 10                                                                    & \texttt{Progression}, \texttt{XOR}, \texttt{OR}, \texttt{AND}, \texttt{consistent union} \\
              & Size           & 10                                                                    & \texttt{Progression}, \texttt{XOR}, \texttt{OR}, \texttt{AND}, \texttt{consistent union}                                            \\
              & Number         & 10                                                                    &      \texttt{Progression}, \texttt{consistent union}                                       \\
              & Position       & 9                                                                   &  \texttt{XOR}, \texttt{OR}, \texttt{AND}                                         \\
              & Type           & 7                                                                     &  \texttt{XOR}, \texttt{OR}, \texttt{AND}, \texttt{consistent union}                                           \\
\cmidrule(r){1-1}\cmidrule(r){2-2}\cmidrule(r){3-3}\cmidrule(r){4-4}
Line          & Color          & 10                                                                    &               \texttt{Progression}, \texttt{XOR}, \texttt{OR}, \texttt{AND}, \texttt{consistent union}                              \\
              & Type       & 6      &    \texttt{XOR}, \texttt{OR}, \texttt{AND}, \texttt{consistent union}        \\
\bottomrule
\end{tabular}
\end{table}

\subsubsection*{\name backend}
Here, we describe the rule probability computation and execution for the PGM dataset. 
Similar to \name's application to the RAVEN dataset, the \texttt{progression} rule is implemented with VSA-enhanced vector operations.
Moreover, the \texttt{consistent union} rule implementation benefits from the VSA-enabled computation-in-superposition, similar to the \texttt{distribute three} rule in RAVEN.
The logical rules (\texttt{XOR}, \texttt{OR}, and \texttt{AND}) are simple logical operations that can be implemented more efficiently in the original low-dimensional PMF space.
We compute the rule probability along the rows and columns and execute it accordingly. 
In the following, we describe the row-wise implementation; the column-wise implementation is done by feeding the transposed context matrix to the \name backend.  
\setcounter{paragraph}{0}
\paragraph{Progression.}
The progression rule's probability computation and execution are implemented as described in equation~\eqref{eq:progression} and equation~\eqref{eq:progression-exe}, respectively, where only the increment by one value is detected and executed in this case. 
\paragraph{Consistent union.}
The consistent union rule relates to a discrete concept; hence, we use random codewords. 
First, we compute the row-wise binding of the PMF-vector representation of the first two rows:
\begin{align}\label{eq:consistentunion}
     \mathbf{r}_{i} = \mathbf{a}^{(i,1)} \odot \mathbf{a}^{(i,2)} \odot \mathbf{a}^{(i,3)}, \quad i\in \lbrace 1,2 \rbrace. 
\end{align}
The rule probability is computed by 
\begin{align}
    \mathbf{u}[\texttt{consistent union}] = \mathrm{sim}(\mathbf{r}_1,\mathbf{r}_2)\cdot h_{c.u.}(\mathbf{a}^{(1,1)},\mathbf{a}^{(1,2)}, ..., \mathbf{a}^{(2,3)}), 
\end{align}
where 

\begin{align}
    h_{c.u.}(\mathbf{a}^{(1,1)},\mathbf{a}^{(1,2)}, ..., \mathbf{a}^{(2,3)}):= \left( \prod_{i\in\{1,2\}} \prod_{j\in\{1,2\}} (1-\mathrm{sim}(\mathbf{a}^{(i,j)},\mathbf{a}^{(i,j+1)})\right) \cdot  (1-\mathrm{sim}(\mathbf{a}^{(3,1)},\mathbf{a}^{(3,2)})
\end{align}
validates the constraint that panels are not equal within a row.
If the rule is selected, it is executed by
\begin{align}
    \mathbf{\hat{a}}^{(3,3)} = \mathbf{r}_1 \circledast \left(\mathbf{a}^{(3,1)} \odot \mathbf{a}^{(3,2)} \right).
\end{align}

\paragraph{XOR, OR, and AND.}
The rule probability of the logical rules is computed in the original PMF space by summing up all possible rule implementations. 
For example, the rule probability for the \texttt{XOR} rule is determined by: 
\begin{align}
u[\texttt{XOR}]= \left(\sum_{r=1}^2  \sum_{\substack{v_1, v_2, v_3\\  s.t.\, 1_{XOR} (v_1, v_2, v_3)}}   \prod\limits_{c=1}^3 \mathbf{p}^{(r,c)}[v_c]\right) + \sum\limits_{\substack{v_1, v_2}}   \prod\limits_{c=1}^2 \mathbf{p}^{(3,c)}[v_c], 
\end{align}
where $1_{XOR}(v_1, v_2, v_3)$ indicates the correctness of the \texttt{XOR} rule within a row/column given the indices $v_1$, $v_2$, and $v_3$. 
Similarly, the rule probability for \texttt{OR} and \texttt{AND} are determined with the corresponding indication function. 
For executing the rule, we marginalize all combinations in the last row, i.e., 
\begin{align}
    \mathbf{\hat{p}}^{(3,3)}[v] = \sum_{\substack{v_1, v_2\\  s.t.\, 1_{XOR} (v_1, v_2, v)}} \prod\limits_{c=1}^2 \mathbf{p}^{(3,c)}[v_c]. 
\end{align}

\begin{table}[]
\centering
\caption{End-to-end accuracy (\%) on the neutral split of the PGM test set. The upper part shows the results reported in the literature. In the lower part, we report the average accuracy $\pm$ the standard deviation over five runs with different seeds for our \name and the reproduced baselines. }
\label{tab:pgm_results}
\begin{threeparttable}
\begin{tabular}{ll}
\toprule
Method      & Accuracy \\
\cmidrule(r){1-1}\cmidrule(r){2-2} 
CNN+MLP~\cite{barrett2018measuring}     & 33.0     \\
LSTM~\cite{barrett2018measuring}        & 35.8     \\
Resnet-50~\cite{barrett2018measuring}   & 42.0     \\
Wild-ResNet~\cite{barrett2018measuring} & 48.0      \\
CoPINet~\cite{CoPINET_19}               & 56.4     \\
WReN~\cite{barrett2018measuring}        & 62.8     \\
LEN~\cite{zheng2019abstract}            & 88.9    \\
SCL~\cite{wu2020scl}                    & 88.9    \\
MRNet~\cite{MRNet_CVPR2021}             & 93.4  \\    
\cmidrule(r){1-1}\cmidrule(r){2-2} 
PrAE~\cite{PrAE_CVPR21}                 & N/A$^{\star}$ \\
LEN~\cite{zheng2019abstract}            & N/A$^{\dagger}$    \\
SCL~\cite{wu2020scl}                    & N/A$^{\dagger}$     \\
MRNet~\cite{MRNet_CVPR2021}             & 68.34$\pm$4.73  \\    
NVSA (end-to-end tr.)                   & 68.30$\pm$4.93\\
\bottomrule
\end{tabular}
\begin{tablenotes}
\footnotesize
\item[*] PrAE only applied to RAVEN.
\item[$\dagger$] No code for PGM available on the code repositories. 
\end{tablenotes}
\end{threeparttable}
\end{table}

\subsubsection*{Selection of the final answer}
For every attribute, we execute the rule with the highest probability yielding the estimated PMFs for the line and shape constellation: $\hat{P}:= \{ \hat{P}_L,\hat{P}_S\}$. 
Finally, we compute the score of every candidate answer panel $i$ by summing up the JSD of the individual attributes: 
\begin{align}
    s(P^{(i)},\hat{P}) = - \sum_a w_a \cdot \mathrm{JSD}(\mathbf{p}^{(i)}_a, \mathbf{\hat{p}}_a), 
\end{align}
where $w_a$ weights the contribution of the attribute $a$. 
In the RAVEN dataset, all the attributes equally contribute to the final score (i.e., $w_a=1, \, \forall a$). 
In contrast, in the PGM dataset, only 1 to 4 attributes have an active rule. 
Hence, we use a learnable small-sized multi-layer perceptron (MLP) which predicts the set of active attributes given the rule probabilities and the JSD errors. 
More concretely, the MLP takes the concatenation of all rule probabilities, their maximizing values, and the JSD errors as input and predicts the values $w_a$. 
The MLP contains one hidden layer with a dimension of 75 and a ReLU activation, and a sigmoid activation at the output. 
The MLP is learned by optimizing the binary cross-entropy loss between the predicted attribute weights and the ground-truth values, which are derived from the auxiliary attribute rules.
\subsubsection*{Training details}
We train each perception frontend separately with training data containing examples with the corresponding attribute rules. 
For example, for training the frontend corresponding to the position and type in the shape constellation, we filter the training set such that it contains either rules with attribute position or type. 
For improved training, we restrict the training samples to have only one rule. 
This yields around 314,000 examples for training the type-color frontend for the line constellation and 146,000 examples for each shape constellation frontend (position-color, position-type, and position-size). 
The models are trained and validated on a Linux machine using an NVIDIA Tesla A100 GPU.

Similar to our approach on RAVEN, we optimize the REINFORCE loss, which is augmented with an auxiliary loss (see equation~(19) in Methods), where only attributes with active rules contribute to the loss (provided by meta-labels). 
We train the shape-related frontends for 45 epochs and the line-related one for 25 epochs using the Adam optimizer with weight decay $10^{-4}$, a constant learning rate of $9.5\times 10^{-5}$, and batchsize of 16. 

After the \name training, the attribute selection MLP is learned on a randomly selected subset of the complete training data (i.e., no rule-based filtering), which turned out to be sufficient to the large dataset size.  
We train the MLP for 50 epochs using a batchsize of 64 and a learning rate of 0.01, where we randomly select only 128 batches in each epoch (0.68\% of the entire dataset).

\subsubsection*{Experimental results}
Supplementary Table~\ref{tab:pgm_results} compares the end-to-end accuracy of our \name with various baselines. 
The upper part of the table shows the accuracy reported in the literature, where the highly accurate methods are LEN~\cite{zheng2019abstract} (88.9\%), SCL~\cite{wu2020scl} (88.9\%), and MRNet~\cite{MRNet_CVPR2021} (93.4\%). 
The lower part of the table compares the accuracy of the reproduced methods with our \name.  
PrAE~\cite{PrAE_CVPR21} was only developed for the RAVEN dataset; hence, it could not be easily applied to the PGM dataset.  
Similarly, the open-sourced code of LEN~\cite{zheng2019abstract} and SCL~\cite{wu2020scl} can only be applied to RAVEN, even though PGM results are reported in the corresponding works.
Finally, MRNet~\cite{MRNet_CVPR2021}, the current state-of-the-art method on PGM, provides code for this dataset. 
However, training the architectures from scratch with randomly initialized weights with different seeds yielded significantly lower accuracy than the one reported in their paper (68.3\% vs. 93.4\%), despite optimizing the weight decay for better training. 
Our \name achieves an average accuracy of 68.3\%, being competitively with the reproduced MRNet.

\end{document}